%% file: ecai2025.tex
\pdfoutput=1
\documentclass{ecai} 

\usepackage{natbib}

\usepackage{times}
\usepackage{soul}
\usepackage{url}
\usepackage[hidelinks]{hyperref}
\usepackage[utf8]{inputenc}
\usepackage[small]{caption}
\usepackage[inline]{enumitem}
\usepackage{array}
\usepackage{graphicx}
\usepackage{subcaption} %required for subfigures
\usepackage[table,xcdraw]{xcolor}
\usepackage{comment}
\usepackage{amsmath}
\usepackage{amsthm}
\usepackage{booktabs}
\usepackage{algorithm}
\usepackage{algorithmic}
\usepackage[switch]{lineno}
\usepackage{makecell} % To allow line breaks in table cells
\usepackage{multirow}
\usepackage{adjustbox} % For table scaling

\definecolor{googleGreen}{HTML}{B6D7A8} 
\definecolor{googleBlue}{HTML}{A4C2F4} 
\definecolor{googlePurple}{HTML}{B4A7D6}

\definecolor{googleBlueDarker}{HTML}{3C78D8}
\definecolor{googleGreenDarker}{HTML}{38761D}
\definecolor{googlePurpleDarker}{HTML}{674EA7}

\newcommand{\BibTeX}{B\kern-.05em{\sc i\kern-.025em b}\kern-.08em\TeX}

\begin{document}

%%%%%%%%%%%%%%%%%%%%%%%%%%%%%%%%%%%%%%%%%%%%%%%%%%%%%%%%%%%%%%%%%%%%%%%%

\begin{frontmatter}

%%% Use this command to specify your submission number.
%%% In doubleblind mode, it will be printed on the first page.

\paperid{7867} 

%%% Use this command to specify the title of your paper.

\title{Mapping User Trust in Vision Language Models: \\Research Landscape, Challenges, and Prospects}

%%% Use this combinations of commands to specify all authors of your 
%%% paper. Use \fnms{} and \snm{} to indicate everyone's first names 
%%% and surname. This will help the publisher with indexing the 
%%% proceedings. Please use a reasonable approximation in case your 
%%% name does not neatly split into "first names" and "surname".
%%% Specifying your ORCID digital identifier is optional. 
%%% Use the \thanks{} command to indicate one or more corresponding 
%%% authors and their email address(es). If so desired, you can specify
%%% author contributions using the \footnote{} command.

\author[A]{\fnms{Agnese}~\snm{Chiatti}\thanks{Corresponding Author. Email: agnese.chiatti@polimi.it}}
\author[B]{\fnms{Sara}~\snm{Bernardini}}
\author[C]{\fnms{Lara}~\snm{Shibelski Godoy Piccolo}} 
\author[A]{\fnms{Viola}~\snm{Schiaffonati}}
\author[A]{\fnms{Matteo}~\snm{Matteucci}}

\address[A]{Politecnico di Milano, Italy}
\address[B]{University of Oxford, UK}
\address[C]{CODE University of Applied Sciences, Germany}

%%% Use this environment to include an abstract of your paper.

\begin{abstract}
The rapid adoption of Vision Language Models (VLMs), pre-trained on large image-text and video-text datasets, calls for protecting and informing users about when to trust these systems. This survey reviews studies on trust dynamics in user-VLM interactions, through a multi-disciplinary taxonomy encompassing different cognitive science capabilities, collaboration modes, and agent behaviours. Literature insights and findings from a workshop with prospective VLM users inform preliminary requirements for future VLM trust studies.
\end{abstract}

\end{frontmatter}

%%%%%%%%%%%%%%%%%%%%%%%%%%%%%%%%%%%%%%%%%%%%%%%%%%%%%%%%%%%%%%%%%%%%%%%%
\section{Introduction}
Vision Language Models (VLMs) have emerged as a methodological shift for learning correspondences between image-text and video-text pairs from large web-scale data. Unlike traditional Computer Vision methods, VLMs reduce the dependency on curated, task-specific training datasets, enabling zero-shot inference on unseen tasks and categories without additional training \cite{zhang2024vision}. 
Thanks to their remarkable ability to interpret image and video content, these models are being rapidly adopted by society at large. This unprecedented adoption rate opens up exciting opportunities, but it also raises significant concerns. Indeed, VLMs are inherently challenging to audit, often closed-source, or function as black-box systems that can be analysed only via indirect observation, steering Artificial Intelligence (AI)  research closer to an ``ersatz natural science” \cite{kambhampati2024can}. 

These challenges become especially pronounced when thinking of VLMs as components of complex robotic systems deployed in safety-critical environments, where errors in decision-making can lead to catastrophic consequences - %\cite{cummings2021rethinking} - 
like in the case of autonomous transportation, %\cite{perez2024artificial}
inspection and maintenance, %\cite{jovan2023adaptive}
disaster response, %\cite{schiaffonati2016stretching}
and assistive healthcare. %\cite{chiatti2023visual}
%The unprecedented rate of adoption of AI tools by society at large makes this problem extremely nuanced, limiting our ability to grasp the broader impact of this technology and all the ramifications of its design, development, end uses and misuses \cite{grote2024taming}. 
Human oversight in assessing VLM performance within real-world settings is essential to safeguard users about how and when to trust these systems. 

The EU Ethics Guidelines for Trustworthy AI \cite{EUTAI} and AI Act \cite{EUAct} offer a regulatory framework for the trustworthiness of AI based on ethical principles, varying levels of risk, and technical requirements. However, defining Trustworthy AI (TAI) within specific real-world contexts remains a topic of active debate \cite{zanotti2023keep}. The first gap is epistemological, requiring a distinction between \textit{trustworthiness}, an inherent property of a system's actual capabilities, and \textit{trust}, which reflects the user's perception of trustworthiness \cite{mayer1995abi}. Moreover, the term \textit{trust} has been borrowed from interpersonal relationships and directly applied to AI. However, further research and regulatory efforts are needed to map this notion appropriately in the context of human-AI interactions. 

Adding to the complexity, the concept of trust overlaps with other dimensions that characterise Human-AI interaction, such as \textit{explainability} \cite{chander2024xai}, \textit{transparency} \cite{barman2024beyond}, \textit{fairness} \cite{calegari2023assessing}, and \textit{accountability} \cite{novelli2024accountability}, to name just a few. Thus, the lack of consensus around user trust in AI stems, at least in part, also from a lack of clarity on the different socio-technical factors contributing to trust building and the interdependencies between these factors. \\
\textbf{Problem scope.} Trust dynamics are hard to define because they are subjective and contextual to the considered application scenario. For instance, we would likely place a different level of trust in a VLM if we asked it to describe a video intended for entertainment versus one used for medical diagnosis. Ultimately, perceived trust evolves over time: it is influenced by our prior experience and familiarity with the technology, %\cite{rosen_previous_2024}, 
and it develops over successive interactions \cite{mehrotra2024integrity}. Given the broad and multi-faceted nature of the field of TAI, our focus is on \textit{reviewing current efforts in studying trust dynamics during the interaction between users and Vision Language Models}.      \\
\textbf{Novelty and contributions.} Recent surveys have examined various factors contributing to user trust in AI: i) explainability \cite{chander2024xai}, ii) model and data fairness \cite{calegari2023assessing}, iii) robustness \cite{chander2024xai,tocchetti2024robustness}, along with trade-offs between fairness, robustness, and iv) model accuracy \cite{li2024triangular}. Some studies focus on trust across the AI lifecycle \cite{calegari2023assessing}, in software development \cite{liu2024tertiary}, or applications in autonomous robotics and safety-critical domains \cite{methnani2024who,perez2024artificial}. %, with causality as a key factor for adequate explanations \cite{rawal2024causality},
While these surveys provide valuable overviews independent of specific AI models and data modalities, fewer studies have explored trust and trustworthiness within Computer Vision. %For example, \cite{wan2024data} review efforts to refine biased or untrustworthy Computer Vision datasets. 
A growing number of frameworks, such as the comprehensive TrustLLM framework \cite{huang2024trustllm}, have been recently proposed to conceptualize trust in the context of Large Language Models (LLMs). However, advancing the inquiry to Vision Language Models necessitates additional consideration of how visual perception and multimodal reasoning influence trust formation.
We share a focus on VLMs with the survey by Liu et al. \cite{ijcai2024liu_vlm}, which compares safety features like robustness against attacks and harmful content. In contrast, we review user trust in VLMs across different cognitive capabilities, human-AI collaboration modes, and agent behaviours, offering a multi-disciplinary perspective that prioritizes user engagement as central to TAI studies. 
%%%% Other srveys found but not included above:%%%%%%%%%%%%%%
% \cite{ijcai2024li_graphs} Knowledge graphs & LLM too specific to cite here -not related to TAI- but may be useful to include later on\\
% \cite{lorena2024trusting} about going beyond quality of predictions: partially related to data bias, but more about inadequacy of benchmark metrics, in part. Could be relevant elsewhere. 
%%%%%%%%%%%%%%%%%%%%%%%%%%%%%%%%%%%%%%%%%%%%%%%%%%%%%%%%%%%%%
To this aim, we contribute: 
\begin{itemize}[topsep=0pt, partopsep=0pt]
    \item A \textbf{novel taxonomy} for categorising related work on trust dynamics in user-VLM interaction;
    \item A \textbf{systematic review} of current methods and datasets based on the proposed taxonomy;
    \item \textbf{Insights from a workshop with users} to ground the notion of trust in VLMs in a concrete use case.
\end{itemize}

\textbf{Roadmap.} Section 2 illustrates the literature search protocol and proposed taxonomy for studying trust in human-VLM interaction. Section 3 presents workshop results from prospective VLM users, complementing literature insights with user-driven requirements. Together, findings from the survey and workshop inform the discussion on current gaps and future improvement directions in Section 4.  

\section{Vision Language Models (VLMs)}  
Vision Language Models (VLMs) have emerged as a methodological shift for learning correspondences between image-text (and video-text) pairs retrieved from large web-scale data. Unlike traditional Computer Vision methods, VLMs reduce the dependency on curated, task-specific datasets for training Deep Neural Networks (DNNs). This learning paradigm typically involves three stages. A \textit{pre-training} phase, where the model is optimised using large-scale, off-the-shelf data, either labelled or unlabelled. An optional \textit{fine-tuning} phase, where the model is adapted to domain-specific data. An \textit{inference} phase, where the model performance is evaluated on downstream tasks. Notably, VLMs have gained particular attention for their ability to skip the fine-tuning step, enabling zero-shot inference on unseen tasks or categories without additional training \cite{zhang2024vision}. 

Zero-shot capabilities in VLMs arise from their design, as visual and textual features are extracted via encoder modules in a general-purpose manner, i.e., independently of any specific downstream task. Visual features are typically extracted from images and video frames using either Convolutional Neural Networks (CNNs) or Transformers, while textual features are almost invariably extracted with Transformers. In the case of video inputs, features are extracted frame-by-frame using CNNs or Transformers and then aggregated. 

Correlations between visual and textual features are learned through various pre-training objectives. These include i) \textit{contrastive objectives}, which optimise embeddings for positioning similar features closer and dissimilar features farther in the vector space; %\cite{yang2022unified}; 
ii) \textit{generative objectives}, where correlations are learned by generating data within a single modality (e.g., image-to-image generation \cite{luo2023segclip}) 
or across modalities - e.g., text-to-image generation \cite{singh2022flava}; iii) \textit{alignment objectives}, which focus on directly matching corresponding elements in visual and textual inputs (e.g., matching local image regions to words \cite{li2022glip}). 
In the case of video inputs, models are often trained on iv) \textit{temporal objectives}, like re-ordering input frames \cite{chen2023vlp}.   

Earlier VLMs used separate branches for visual and textual modalities during pre-training as in the original CLIP model \cite{radford2021learning}. However, recent architectures have shifted to unified designs that employ a single encoder for both modalities \cite{jang2023singletower,singh2022flava} to enable cross-modal feature fusion and reduce computational overhead \cite{zhang2024vision}.

\section{Mapping User Trust in VLMs} 
%Methodology adopted for conducting the literature review in a systematic fashion. 

\subsection{Search and selection criteria\label{sec:lit_search}}
We systematised the literature search by year of publication, textual search keywords, and publication venue. 

Dimensions AI, the world's largest linked research information dataset, indexes 271,976 publications featuring the keyphrase \textit{Vision Language Model} in the past year alone. To account for the fast proliferation of works on VLMs and prioritise newly-emerging trends, we limited the search to works published in the last year %between Jan 2024 and Jan 2025 
(though we also considered datasets released earlier cited in these works). We used combinations of the search keywords \textit{trustworthy/trustworthiness}, \textit{user study}, \textit{human in the loop}, \textit{vision}, \textit{language}, \textit{(user) trust evaluation}, \textit{evaluation method}, \textit{interaction}, along with common acronyms for large-scale models: Large Language Model (\textit{LLM}), Vision Language Model (\textit{VLM}), Large Multimodal Model (\textit{LMM}), Multimodal Large Language Model (\textit{MLLM}). We supplemented Google Scholar results with checks on a multidisciplinary range of journals (Artificial Intelligence, Transactions on Machine Learning Research, Pattern Recognition Letters, Int. Journal of Social Robotics, Int. Journal of Human-Computer Interaction, ACM Computing Surveys Special Issue on TAI) and conferences (IJCAI, ECAI, AAAI, NeurIPS, CVPR, ICML, ICLR, CoRL, ICRA, IROS, CHI, HHAI, FAccT). 

This protocol yielded 157 candidate papers. After screening the abstracts, we filtered out works nor centred around VLMs nor examining trust properties. Ultimately, 43 papers were retained to be included in this survey. %: i.e., only X\% of identified papers on AI trust concern visual perception.  
%identified a set of target papers and repeated the search criteria on their citing papers. We 

\subsection{Proposed taxonomy\label{sec:taxonomy}}
% Initially based on pragmatic requirements of: i) focusing on Computer Vision tasks, ii) considering trustworthiness metrics as opposed to only evaluating on accuracy and capabilities/reliability, iii) adopting a human-in-the-loop, human-centric perspective and conducting studies involving actual users of the system.
%Now expanded to be more general-purpose and of interest to a wider audience
A seminal framework for modelling trust within organisations is the ABI framework by Mayer \cite{mayer1995abi}. In the ABI model, trust is defined as a relation between a trusting party (\textit{trustor}) and a counterpart to be trusted (\textit{trustee}), based on the trustor's belief in the trustee's Ability, Benevolence, and Integrity. \textit{Ability} refers to the skills and competencies required to perform a task. \textit{Benevolence} reflects a willingness to act in the best interest of a specific trustor. \textit{Integrity} relates to the trustor's perception that the trustee upholds acceptable principles. Thus, while trustworthiness can be seen as a characteristic inherent to the trustee, trust is the perceived trustworthiness of the trustee from the trustor's perspective.  

\begin{figure}[t!]
    \centering
    \includegraphics[width=\columnwidth, trim={15 25 353 15}, clip]{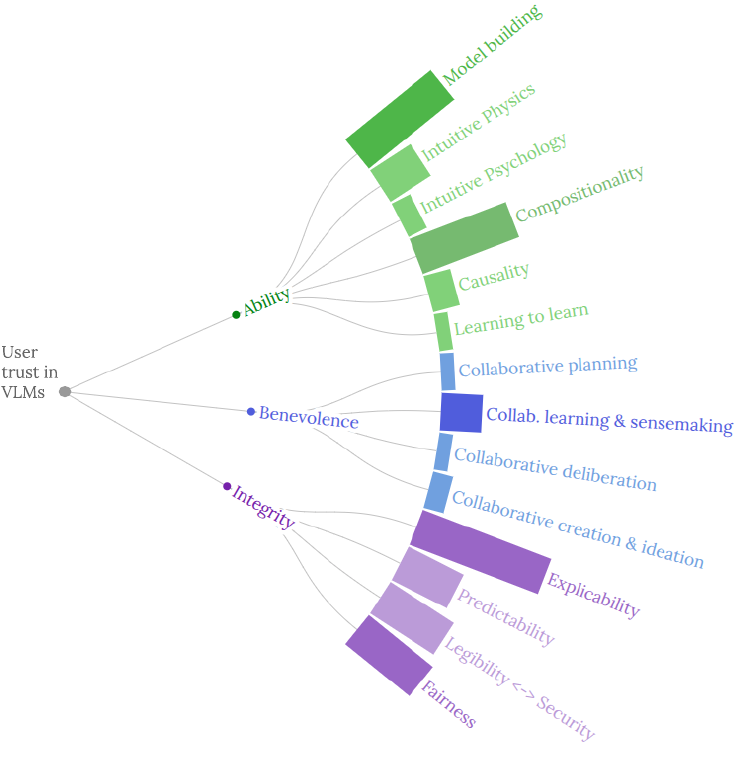} 
    \caption{Proposed taxonomy for modelling user trust in Vision Language Models. The size of leaf nodes is proportional to the count of surveyed papers for each category.}
    \vspace*{3ex}
    \label{fig:taxonomy}
\end{figure}

On the one hand, these trust factors were designed for interpersonal relationships and must be adapted for human-AI collaboration. On the other hand, measuring trustworthiness only in terms of performance-driven metrics and technical requirements without recognizing that trust is the user-perceived trustworthiness of a system overlooks the human element of trust building.   
To address both aspects and ground them in the context of mixed human-VLM teams that cooperate towards shared goals, we extend the ABI framework through key concepts in Situated Cognition \cite{collins2024building,lake2017building} and Theory of Mind \cite{verma2024tom}, proposing the taxonomy in Figure  \ref{fig:taxonomy}. Indeed, we argue that trust in VLMs is built through the collaboration between users and artificial agents that exhibit human-like visual intelligence and comply with cooperation and integrity principles. 

Several recent frameworks aim to categorize trust and trustworthiness in the context of LLM adoption, with the TrustLLM framework \cite{huang2024trustllm} serving as a particularly comprehensive example. However, extending the investigation to VLMs requires also to consider the components of visual perception and reasoning that contribute to trust building.  

To account for this requirement, we begin by characterising VLMs {\color{googleGreenDarker}\textbf{Ability}} through the core ingredients of intelligent machines originally introduced by Lake et al. \cite{lake2017building}, also linked to the cognitive science motifs proposed in Collins et al. \cite{collins2024building}, which we contextualise in the domain of visual perception. 
\begin{description}[leftmargin=0.7em]
    \item[Learning as model building.] In Situated Cognition, knowledge and thinking are necessarily contextual to the environment in which they occur. Thus, human-like learning occurs through building mental models of the world, including environmental models acquired through visual perception. 
    \item[Intuitive Physics] refers to fundamental Physics principles already understood and applied by children early in development, such as object persistence, cohesion, and solidity - allowing them, e.g., to localise objects even when they disappear from their field of view. 
    \item[Intuitive Psychology.]Infants also understand that animate agents act in a goal-oriented and socially-sensitive manner. These skills allow us to infer others' beliefs, intentions, desires, and emotions, even just from visual cues. 
    \item[Causality.] In addition to physics and psychological models, another core capability of human-like intelligence is that of building causal models of the world, e.g., reasoning on cause-effect chains of events represented in a video. 
    \item[Compositionality] is the capability of constructing new representations from building-block components. In Computer Vision, it relates to comprehending image/video constituents, such as object parts and spatial relations.  
    \item[Meta-learning] refers to the capability of ``learning to learn", in addition to continuously learning new concepts and skills. In Computer Vision, it falls under the objectives of Domain Adaptation and Continual Learning. 
    %\item[Thinking fast.] Despite building rich and dense world models, human (visual) perception and thought operate at speed.  
\end{description}
 
Second, to categorise {\color{googleBlueDarker}\textbf{Benevolence}} in the context of human-VLM cooperation, we rely on collaborative thought modes in Collins et al. \cite{collins2024building} that frame AI models as constructive ``thought partners" for users.
\begin{description}[leftmargin=0.7em]
    \item[Collaborative planning] enables joint decision-making and mutual assistance on specific goals and tasks like in the case of service and assistive robots. 
    \item[Collaborative learning and sensemaking] entails pair/team problem solving, the reciprocal identification of knowledge gaps and the joint construction of new problems, e.g., when VLMs are used as educational tools. 
    \item[Collaborative deliberation] contributes to consensus formation through a process of debate, argumentation, critical review, and discussion. 
    %\item[Collaborative sensemaking] can be achieved via explanation and visualization, as well as by analysing data together. 
    \item[Collaborative creation and ideation] refers to co-designing, brainstorming, and critiquing ideas within mixed human-AI teams, like in the case of machine-assisted writing and collaborative sketching.    
\end{description}

Lastly, we build on recent work \cite{verma2024tom} to describe {\color{googlePurpleDarker}\textbf{Integrity}} through agent behaviour types, to which we add the fairness principle, already well-studied in Natural Language Processing \cite{huang2024trustllm}. 

\begin{description}[leftmargin=0.7em]
\item[Explicability] assesses how well a plan aligns with the observer's expectations, given a specific goal or model output (e.g., given a robot-generated plan of actions). 

\item[Predictability] minimizes uncertainty about potential plans associated with a given goal, output, or explanation. 

\item[Legibility $\leftrightarrow$ Security.] \textbf{Legibility} reduces the ambiguity over the agent's goals and intents that can be inferred from its observable behaviour. The counterpart to legible behaviours are obfuscation behaviours. Thus, depending on the context, legibility requirements may need to be traded off with increasing \textbf{security} against adversarial attacks.  

%\item[Security v. obfuscation.] Goals or plans (outputs or explanations) are obfuscated if they do not reveal the true intentions of the agent, even in the absence of active deceit.  

\item[Fairness] involves avoiding biased or discriminatory outcomes (plans, explanations) and ensuring that all users and groups are treated equitably.  
\end{description}

\renewcommand{\arraystretch}{1.2} % Adjust row height
\setlength{\tabcolsep}{3pt} % Reduce horizontal padding in cells
% Define custom column type for wrapped text
\newcolumntype{P}[1]{>{\centering\arraybackslash}p{#1}}

\begin{table*}[ht]
\centering
\caption{Mapping of reviewed methods and datasets to the proposed taxonomy.}
\small % Reduce font size
\begin{adjustbox}{max width=\textwidth} % Scale to page width
\begin{tabular}{|P{2cm}|*{14}{P{2cm}|}} % Adjust column widths
\hline
\multirow{2}{*}{\cellcolor{lightgray}} & %\multicolumn{1}{|c|}{} 
\multicolumn{6}{c|}{\cellcolor{googleGreen}\textbf{Ability}} & 
\multicolumn{4}{c|}{\cellcolor{googleBlue}\textbf{Benevolence}} & 
\multicolumn{4}{c|}{\cellcolor{googlePurple}\textbf{Integrity}} \\
\cline{2-15} 
\cellcolor{lightgray} & 
\rotatebox{90}{\makecell[tl]{\textbf{Model} \\ \textbf{building}}} &
\rotatebox{90}{\makecell[tl]{\textbf{Intuitive} \\ \textbf{Physics}}} &
\rotatebox{90}{\makecell[tl]{ \textbf{Intuitive} \\ \textbf{Psychology}}} &
\rotatebox{90}{\textbf{Compositionality}} &
\rotatebox{90}{\textbf{Causality}} &
\rotatebox{90}{\makecell[tl]{\textbf{Learning} \\\textbf{to learn}}} &
%\rotatebox{90}{\textbf{Thinking fast}} &
\rotatebox{90}{\makecell[tl]{\textbf{Collaborative} \\\textbf{planning}}} &
\rotatebox{90}{\makecell[tl]{\textbf{Collab. learning} \\\textbf{\& sensemaking}}}&
\rotatebox{90}{\makecell[tl]{\textbf{Collaborative} \\\textbf{deliberation}}} &
\rotatebox{90}{\makecell[tl]{\textbf{Collab. creation} \\\textbf{\& ideation}}} &
\rotatebox{90}{\textbf{Explicability}} &
\rotatebox{90}{\textbf{Predictability}} &
\rotatebox{90}{\makecell[tl]{\textbf{Legibility $\leftrightarrow$} \\\textbf{ Security}}} &
\rotatebox{90}{\textbf{Fairness}} \\
\hline
\textbf{Methods \& studies} & \cite{dona2024evaluating},
\cite{xiao2024can},
Pelican \cite{sahu2024pelican},
PIVOT \cite{nasiriany2024pivot}, RoboPoint \cite{yuan2024robopoint} & 
PIVOT \cite{nasiriany2024pivot}, RoboPoint \cite{yuan2024robopoint} & ContextCam \cite{fan_contextcam_2024} & Spatial VLM \cite{chen2024spatialvlm},
Pelican \cite{sahu2024pelican},
PIVOT \cite{nasiriany2024pivot},
RoboPoint \cite{yuan2024robopoint},
VisCoT \cite{shao_viscot_2024},
\cite{zheng_iterated_2024} & 
& & \cite{mehrotra2024integrity} & RLAIF-V \cite{yu2024rlaif}, \cite{mehrotra2024integrity} & PIVOT \cite{nasiriany2024pivot} & ContextCam \cite{fan_contextcam_2024} & \cite{vatsa2024adventures},
RLAIF-V \cite{yu2024rlaif}, 
Pelican \cite{sahu2024pelican},
\cite{mehrotra2024integrity},
DRESS \cite{chen2024dress},
ContextCam \cite{fan_contextcam_2024},
\cite{wang_probabilistic_2024},
OPERA \cite{huang_opera_2024},
VCD \cite{leng_mitigatingVCD_2024},
\cite{gui_conformal_2024},
CGD \cite{deng_seeing_2024},
\cite{fang_uncertainty_2024} & 
\cite{dona2024evaluating}, Pelican \cite{sahu2024pelican}, \cite{mehrotra2024integrity}, 
\cite{wang_probabilistic_2024},
\cite{khan_consistency_2024},
\cite{zheng_iterated_2024}
 & \cite{vatsa2024adventures},
\cite{liu2024safety},
ECSO \cite{gou2024eyes},
\cite{mehrotra2024integrity},
Shadowcast \cite{xu_shadowcast_2024},
\cite{khan_consistency_2024},
\cite{islam_malicious_2024} & \cite{vatsa2024adventures},
\cite{wu2024evaluating},
\cite{liu2024safety},
ECSO \cite{gou2024eyes},
\cite{mehrotra2024integrity},
DRESS \cite{chen2024dress},
FairCLIP \cite{luo_fairclip_2024}\\
\hline
\textbf{Datasets \& benchmarks} & Next-GQA \cite{xiao2024can}, MERLIM \cite{villa2023merlim},
STAR \cite{wu2021star},
AGQA \cite{grunde2021agqa},
RexTime \cite{chen2024rextime},
RoboVQA \cite{sermanet2024robovqa}
,
VHELM \cite{lee_vhelm_2024},
SpatialEval \cite{wang_SpatialEval_2024},
ConvBench \cite{liu_convbench_2024},
ProDG \cite{shirai2024prodg},
ManipVQA \cite{huang_manipvqa_2024}  & 
ReXTime \cite{chen2024rextime}
AGQA \cite{grunde2021agqa},
RoboVQA \cite{sermanet2024robovqa},
SpatialRGPT \cite{cheng2024spatialrgpt},
ManipVQA \cite{huang_manipvqa_2024} & 
ReXTime \cite{chen2024rextime}
STAR \cite{wu2021star} 
& MERLIM \cite{villa2023merlim},
STAR \cite{wu2021star},
AGQA \cite{grunde2021agqa},
Visual CoT \cite{shao_viscot_2024},
SpatialEval \cite{wang_SpatialEval_2024},
ConvBench \cite{liu_convbench_2024},
SpatialRGPT \cite{cheng2024spatialrgpt},
PROVE \cite{prabhu2024trust},
ManipVQA \cite{huang_manipvqa_2024}  &
ReXTime \cite{chen2024rextime}
STAR \cite{wu2021star},
RoboVQA \cite{sermanet2024robovqa},
VHELM \cite{lee_vhelm_2024}
&
STAR \cite{wu2021star},
ConvBench \cite{liu_convbench_2024}&
RoboVQA \cite{sermanet2024robovqa}&
MLLM-as-a-Judge \cite{chen2024MLLMasjudge},
RoboVQA \cite{sermanet2024robovqa},
VLSafe \cite{chen2024dress},
ConvBench \cite{liu_convbench_2024}& 
VLSafe \cite{chen2024dress}&
ConvBench \cite{liu_convbench_2024},
BEAF \cite{ye-bin_beaf_2025}& 
Next-GQA \cite{xiao2024can},
MERLIM \cite{villa2023merlim},
MLLM-as-a-Judge \cite{chen2024MLLMasjudge},
SpatialEval \cite{wang_SpatialEval_2024},
MIC \cite{zhao2024mmicl},
ProDG \cite{shirai2024prodg},
PROVE \cite{prabhu2024trust},
BEAF \cite{ye-bin_beaf_2025} &
MLLM-as-a-Judge \cite{chen2024MLLMasjudge},
VHELM \cite{lee_vhelm_2024},
BEAF \cite{ye-bin_beaf_2025}&
SPA-VL \cite{zhang2024spa}, 
VLSafe \cite{chen2024dress},
PROVE \cite{prabhu2024trust},
BEAF \cite{ye-bin_beaf_2025}&
SPA-VL \cite{zhang2024spa}, 
VLSafe \cite{chen2024dress},
VHELM \cite{lee_vhelm_2024},
Harvard-FairVLMed \cite{luo_fairclip_2024},
SocialCounterfactuals \cite{howard_socialcounterfactuals_2024}\\
\hline
\end{tabular}
\end{adjustbox}
\label{tab:coverage}
\end{table*}

\subsection{Coverage study \label{sec:lit_mapping}}
 The proposed taxonomy allows us to evaluate the current coverage of trust-building factors from a nuanced viewpoint that spans visual cognition capabilities, collaborative thought modes, and AI behavioural motifs. Paper coverage results, already summarised in Figure \ref{fig:taxonomy}, are also detailed in Table \ref{tab:coverage}. %After defining a reference taxonomy, we aim to map related works selected as described in Section \ref{sec:lit_search} to the key dimensions in Figure \ref{fig:taxonomy}. 
 It is worth noting that, in this mapping, the same paper can cover different categories. 

 \paragraph{Methods and studies.} 
 Related research on VLM trust primarily focuses on assessing {\color{googlePurpleDarker}\textbf{Integrity}} violations and mitigating adverse effects. A significant proportion of studies focuses on model explicability \cite{vatsa2024adventures,mehrotra2024integrity,wang_probabilistic_2024}, with particular attention to reducing \textit{hallucinations} - i.e., the emergence of objects or concepts in VLM responses absent from the input prompt \cite{yu2024rlaif,sahu2024pelican,chen2024dress,huang_opera_2024,leng_mitigatingVCD_2024,deng_seeing_2024,fang_uncertainty_2024,fan_contextcam_2024}. A subset of works on explicability aims at aligning models to human preferences \cite{chen2024dress,gui_conformal_2024}, though some studies leverage pre-trained LLMs to generate preferences, bypassing human annotators. 
 
 Other studies address predictability factors, focusing on improving model output consistency \cite{dona2024evaluating,khan_consistency_2024,sahu2024pelican} and enhancing the robustness to input perturbations, such as textual and visual distractors \cite{wang_probabilistic_2024,zheng_iterated_2024}. Research into model legibility and security mainly investigates vulnerability to adversarial attacks \cite{vatsa2024adventures,liu2024safety,gou2024eyes,khan_consistency_2024,islam_malicious_2024,xu_shadowcast_2024}. 
 Chen et al., \cite{chen2024dress} offer a distinct perspective by exploring the likelihood of models to persuade users into causing harm. 
 In the domain of fairness, studies focus on mitigating social biases in model responses \cite{vatsa2024adventures}, particularly the generation of discriminatory, harmful, or toxic content \cite{wu2024evaluating,liu2024safety,gou2024eyes,chen2024dress,luo_fairclip_2024}. 
 
By contrast, research on the perceived cognitive {\color{googleGreenDarker}\textbf{Ability}} of VLMs, which contributes to trust-building, is comparatively sparse. Existing work in this area primarily evaluates spatial and compositional reasoning \cite{sahu2024pelican,shao_viscot_2024}, especially in Embodied AI and Robotics applications \cite{chen2024spatialvlm,nasiriany2024pivot,yuan2024robopoint}. Some studies investigate visual spatio-temporal grounding capabilities in Visual Question Answering (VQA) \cite{xiao2024can,nasiriany2024pivot,liu_convbench_2024,yuan2024robopoint,zheng_iterated_2024}, instruction following \cite{sahu2024pelican}, and video understanding \cite{dona2024evaluating}. Nonetheless, critical aspects such as the VLM’s ability to comprehend Intuitive Physics principles \cite{nasiriany2024pivot}, exhibit social and psychological reasoning skills \cite{fan_contextcam_2024}, engage in causal reasoning, or demonstrate continuous learning remain largely unexplored. %in trust-related studies.

Notably, only a limited body of work investigates trust through user interaction, resulting in minimal coverage of the diverse user-AI collaboration modes that define the {\color{googleBlueDarker}\textbf{Benevolence}} dimension. Some studies adopt Human-in-the-Loop approaches to iteratively refine predictions (collaborative deliberation) and align models with human feedback (collaborative learning), as exemplified by the PIVOT \cite{nasiriany2024pivot} and RLAIF-V methods \cite{yu2024rlaif}. However, similar to the case of human preference alignment \cite{gui_conformal_2024}, ``Model-in-the-Loop" approaches \cite{khan_consistency_2024}—which rely on large-scale AI models to approximate user behaviour—remain dominant presumably due to the logistical complexity, ethical implications, and cost of user studies. Exceptions include Fan et al. \cite{fan_contextcam_2024} and Mehrotra et al. \cite{mehrotra2024integrity}. Fan et al. explore user-VLM collaboration for image co-creation, while Mehrotra et al. investigate whether users are willing to delegate the task of estimating dish calories from images to a VLM (collaborative planning and sensemaking). In the latter study, the trustworthiness of model explanations is evaluated in terms of honesty, transparency, and fairness, comprehensively addressing all integrity categories in Table \ref{tab:coverage}.     \\
 %\vspace{-0.9em}
 \textbf{Datasets and benchmarks.} The majority of reviewed datasets are designed to assess VLM {\color{googleGreenDarker}\textbf{Ability}}. As in related methods, visual grounding benchmarking efforts remain predominant   \cite{villa2023merlim,wu2021star,grunde2021agqa,chen2024rextime,sermanet2024robovqa,lee_vhelm_2024,wang_SpatialEval_2024,liu_convbench_2024,shirai2024prodg,huang_manipvqa_2024}, including tasks such as detecting \textit{hidden hallucinations}—true positives lacking visual grounding \cite{xiao2024can}. Many datasets focus on compositional reasoning \cite{villa2023merlim,wu2021star,grunde2021agqa,shao_viscot_2024,wang_SpatialEval_2024,liu_convbench_2024,cheng2024spatialrgpt,prabhu2024trust,huang_manipvqa_2024}. Overall, the identified datasets encompass a broader range of cognitive abilities than those explored in related studies. ConvBench \cite{liu_convbench_2024}, for example, introduces a multimodal hierarchy spanning from basic perception to advanced creativity (learning to learn). Benchmarks in robotic and spatial reasoning \cite{sermanet2024robovqa,huang_manipvqa_2024,cheng2024spatialrgpt} evaluate Intuitive Physics, focusing on properties such as object affordances and contact actions. Concurrently, video-based benchmarks emphasize situated reasoning tasks, particularly understanding human intentions (Intuitive Psychology) and analysing cause-effect event chains. This focus is evident in the ReX-Time \cite{chen2024rextime}, VHELM \cite{lee_vhelm_2024}, STAR \cite{wu2021star}, and AGQA datasets \cite{grunde2021agqa}. In this context, scene graphs connecting people, objects, and events emerge as a particularly promising representation. In scene graphs, nodes represent objects (e.g., "person", "table") or attributes (e.g., "red", "wooden"),while edges represent relationships between objects (e.g., "on top of", "holding", "next to"). Scene graphs can thus also be seen as collections of text triples composed by a subject, a predicate, and an object (e.g., "bike near house"). Crucially, this representation format can support semantic concept abstraction for meta-learning \cite{wu2021star} and provide ``means-to-an-end" representations of video narratives \cite{chen2024rextime}.    

Regarding {\color{googlePurpleDarker}\textbf{Integrity}}, the surveyed benchmarks examine VLM explicability in terms of: i) hallucination rates \cite{villa2023merlim,chen2024MLLMasjudge,zhao2024mmicl}; ii) alignment of responses (or robot plans) to human preferences \cite{chen2024MLLMasjudge,shirai2024prodg}; and iii) the influence of different modalities (textual or visual) on model outputs \cite{xiao2024can,zhao2024mmicl,wang_SpatialEval_2024}. Predictability, meanwhile, is often measured by assessing VLM robustness to input perturbations \cite{lee_vhelm_2024,ye-bin_beaf_2025} and output consistency across prompts \cite{chen2024MLLMasjudge}. Prabhu et al. \cite{prabhu2024trust} use scene graphs for model response decomposition and verification, enhancing explicability and legibility.
Additionally, SPA-VL \cite{zhang2024spa} categorizes training data by safety preference alignment, addressing harm categories ranging from privacy violations to human-integrity threats (security) and representational or toxic misuse (fairness). A few other fairness benchmarks also focus on mitigating harmful and toxic content generation \cite{chen2024dress,lee_vhelm_2024,luo_fairclip_2024,howard_socialcounterfactuals_2024}.

By contrast, benchmarks evaluating trust through direct online interaction between users and VLMs remain scarce (see Table \ref{tab:coverage}, {\color{googleBlueDarker}\textbf{Benevolence}}). This limitation extends to the collection of ground-truth data, which increasingly relies on LLM-generated outputs \cite{zhang2024spa,prabhu2024trust,zhao2024mmicl}. Nevertheless, we found a few examples of human-VLM collaboration in the context of training robots via dialogue \cite{sermanet2024robovqa}, capturing human annotation and safety preferences \cite{chen2024MLLMasjudge,chen2024dress}, image inpainting tasks \cite{ye-bin_beaf_2025} and visual co-creative activities \cite{liu_convbench_2024}. 

\vspace{-1em}
\section{User trust in VLMs: a case study} 
Evaluating trust dynamics requires a user-centred perspective. As highlighted in Section \ref{sec:lit_mapping}, the current literature lacks comprehensive studies on user trust in VLMs. These studies are inherently complex, as they require aligning the rapid advancements of technical solutions with participants' subjective understanding of state-of-the-art capabilities and limitations to ensure scientifically rigorous results. A pilot case study on user-VLM trust, albeit preliminary, can offer valuable insights and lessons learned to the research community and serve as a foundation for future studies involving a larger participant base, helping mitigate their complexity, risks, and costs. Furthermore, these preliminary findings can inform the design of a digital tool tailored to support evaluations of user interactions with VLMs. 

To this aim, we organised an exploratory workshop guided by two main research questions. First, we sought to explore \textit{how trust dynamics develop when users engage with a VLM to solve collaborative planning and sensemaking tasks that require situated cognition abilities} (\textbf{RQ1}). Second, we aimed to gather expert feedback on \textit{which features should be prioritized in the design of a Web App to evaluate these trust dynamics in user studies} (\textbf{RQ2}).

\paragraph{Use-case scenario.} Participants with a background in computer science and digital design interact with two AI models: an LLM and a VLM that supports video input. The goal is to assess whether these models can be utilized to train robots to understand observational videos depicting everyday interactions between people and objects. % ---repetition, already evident from the RQs) ---
%Hence, we focus on human-AI collaborative planning and sensemaking tasks requiring situated cognition abilities. %, i.e., the ability to understand situations and make appropriate decisions, based on the knowledge gauged dynamically from the surrounding context \cite{wu2021star}. %While situated reasoning is one of the key prerequisites for deploying autonomous robots in the real world, it remains a significant challenge for VLMs. 

\subsection{Materials and methods}
As a complement to our literature review, we conducted a pilot workshop with experts in Design and Development to begin addressing the evident gap in VLM research involving direct user perspectives. While the scale of this workshop limits the generalizability of our findings, it serves as an exploratory step toward identifying key challenges and opportunities for future user-involved studies. Our aim is not to offer definitive conclusions, but to enrich the survey with expert-grounded insights that foreground the importance of user involvement in this domain. \\
Based on findings from our literature review (Section \ref{sec:lit_mapping}), we took the study by Mehrotra et al. \cite{mehrotra2024integrity} as a reference and modelled trust decisions as users' choices to delegate tasks to an AI model. In this view, appropriate trust is achieved when the user's perceived trust in the system matches the actual trustworthiness of the system. %Given the scarcity of related work on benchmarking trust in VLMs through direct engagement with users, two main research questions emerged, motivating the workshop.  
We structured workshop activities in two parts, each targeting one research question.

\paragraph{Preparatory activities.} Before beginning the two-part practical session, we conducted a 20-minute seminar to outline the broader objectives and context of the study. This session was important for ensuring participants shared a common understanding of: i) Vision Language Models and their distinction from Large Language Models, ii) the importance of assessing user trust in VLMs before deploying them in specific robotic scenarios, and iii) the diverse stakeholder groups to consider in future research. These range, depending on the robotic application, from industry and domain experts to the general public (e.g., in the case of household robots).

\begin{figure}[t]
    \centering
    \includegraphics[width=\columnwidth, trim={20 20 30 10}, clip]{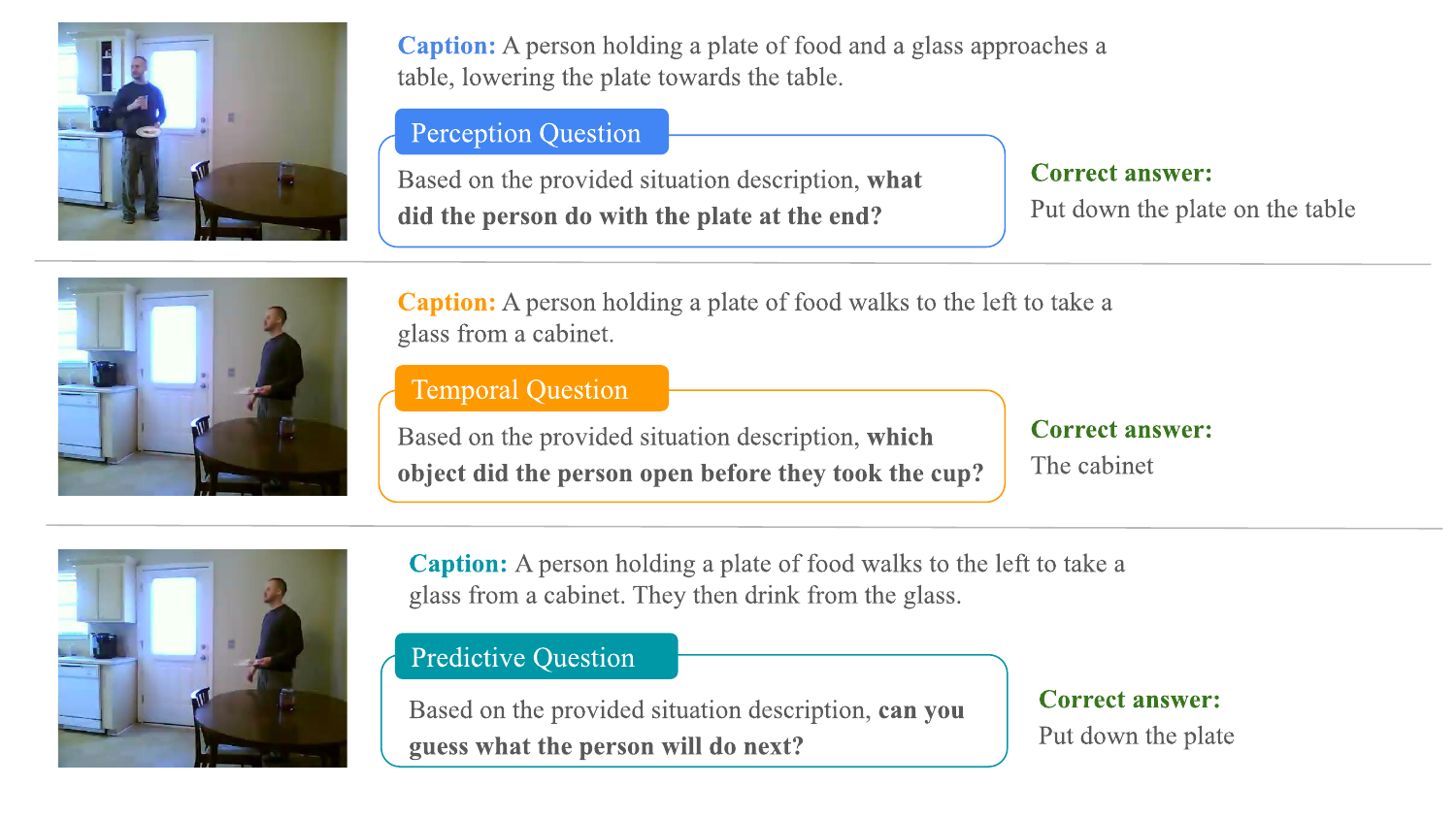} 
    \caption{Question examples within each task, by increasing difficulty from basic perception and model building to advanced abstraction (learning to learn).}
    \label{fig:star_examples}
    \vspace*{3ex}
\end{figure}

\begin{figure*}[t]
    \centering
    % Subfigure 1
    \begin{subfigure}{\textwidth} 
        \centering
    \includegraphics[width=0.4\textwidth,  trim={370 220 0 0}, clip]{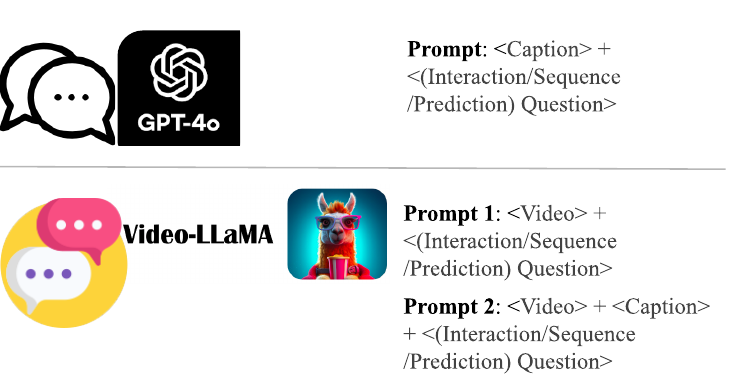} 
    \caption{\textbf{Part 1:} players interact in pairs using different prompts to compare responses from the blindfolded player, the LLM, and the VLM.}
    \label{fig:collab_game}
    \vspace*{3ex}
    \end{subfigure}
    \hfill % Adds space between subfigures
    % Subfigure 2
    \begin{subfigure}{0.45\textwidth} 
         \centering
    \includegraphics[scale=0.45, trim={10 60 30 10}, clip]{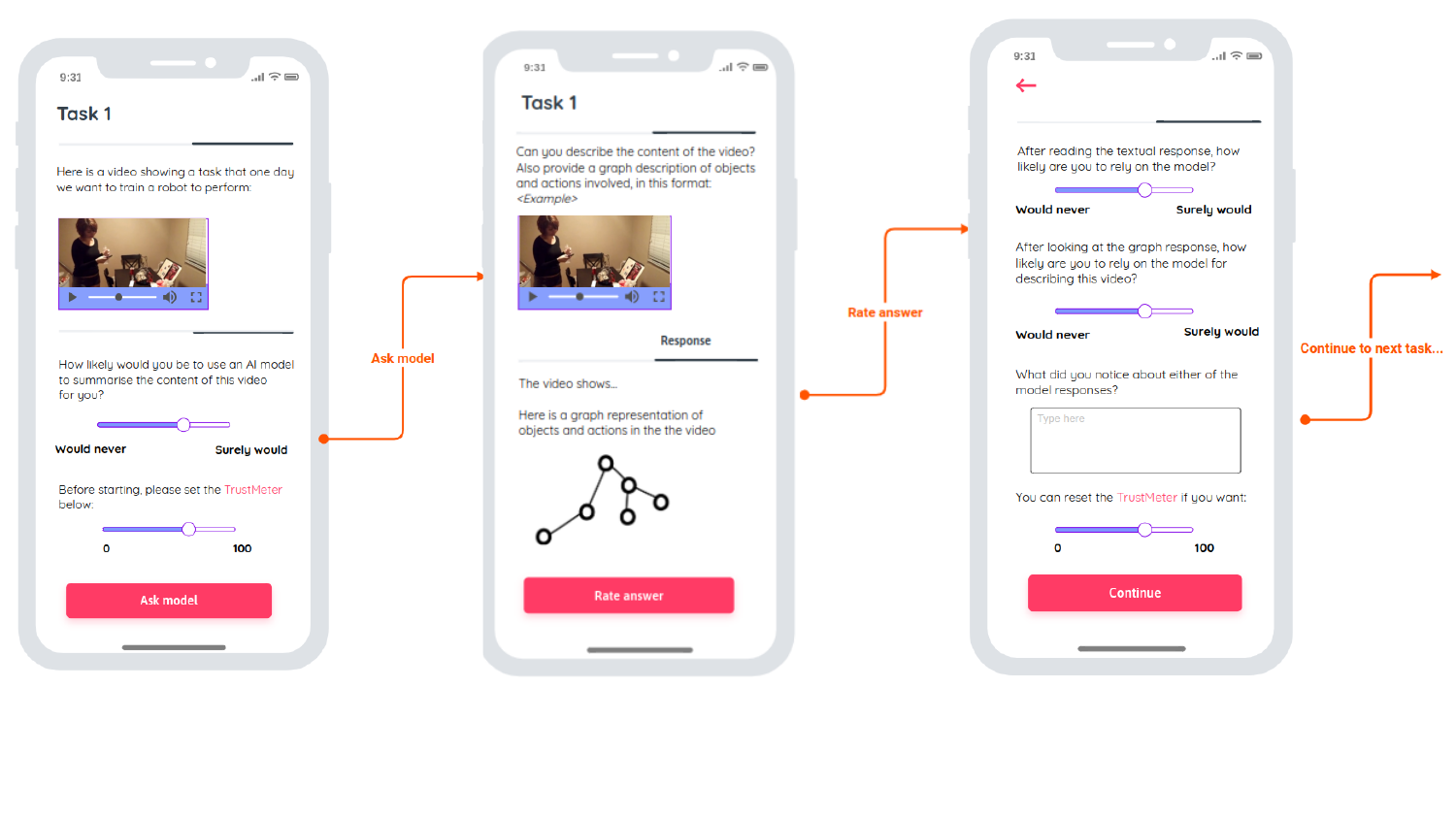} 
    \caption{\textbf{Part 2:} Participants evaluate a set of app mock-ups. }
    \label{fig:mockup}
    \textbf{}
    \end{subfigure}
    \hfill
    \begin{subfigure}{0.45\textwidth} 
         \centering
    \includegraphics[scale=0.40, trim={100 0 200 0}, clip]{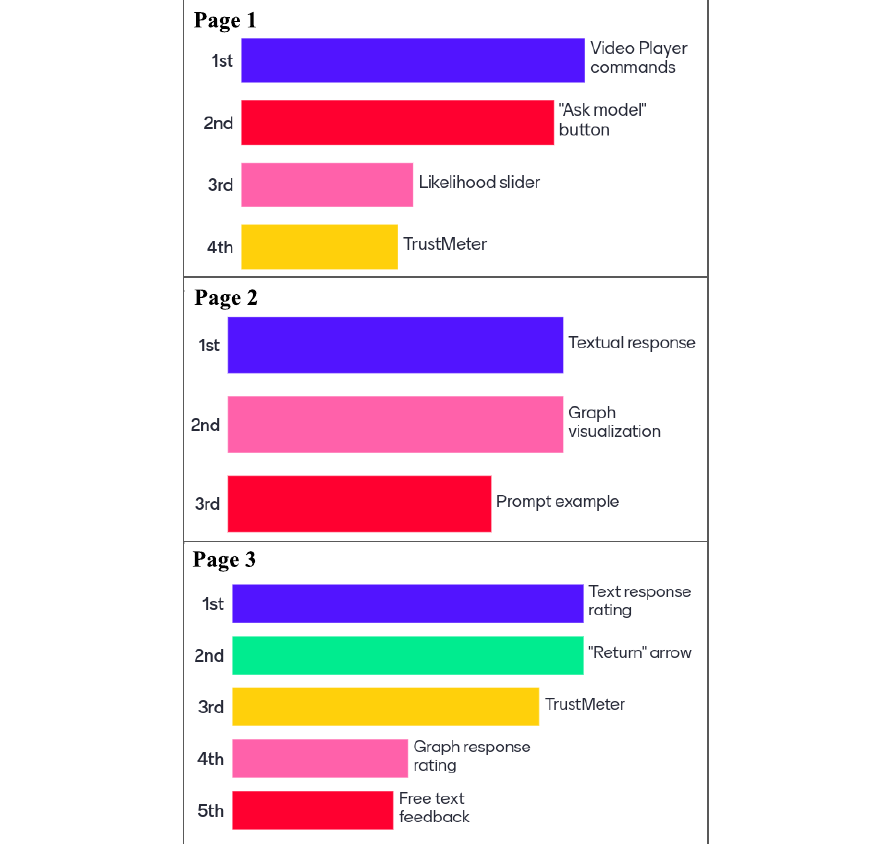} 
    \caption{\textbf{Feature ranking} exercise in part 2. }
    \label{fig:rank}
    \vspace*{3ex}
    \end{subfigure}
    \caption{Structure of the workshop.}
    \label{fig:focus_group}
\end{figure*}

\paragraph{Part 1: Collaborative game.} In the first phase, participants were paired to engage in a collaborative game. One participant was blindfolded, to be led to assess (and trust) only the words of the partner while deprived of other senses. Each  participant ($P1$) interacted with the blindfolded partner ($P2$) and an AI model to solve visual tasks derived from five situations in the STAR benchmark \cite{wu2021star}. Situations consisted of three video segments paired with three questions of increasing difficulty. This choice ensured to cover different levels of VLM cognitive ability \cite{wu2021star,liu_convbench_2024}: from the perception of actions involving people and objects (e.g., \textit{what did a person do...?}), to higher-level reasoning (e.g., \textit{what did the person do before/after X...?}), and forecasting plausible future interactions (e.g., \textit{What will the person do next?}). %an interaction, sequence, and prediction question (Figure \ref{fig:star_examples}). \textbf{Interaction questions} (e.g., \textit{what did a person do...?}) assessed the basic understanding of actions involving people and objects. \textbf{Sequence questions} (e.g., \textit{what did the person do before/after X...?}) evaluated temporal reasoning capabilities. \textbf{Prediction questions} required forecasting plausible future interactions (e.g., \textit{What will the person do next?}). 
 One situation was used as a trial round, and four served as study tasks. Further details on task and question structure are provided in the supplementary material.
 
 We structured the game as follows: $P1$ watched a video, read a textual description aloud to $P2$, and posed a question for $P2$ to answer. $P1$ then compared $P2$’s response to the correct answer and queried ChatGPT4-o using the same textual description. This allowed the comparison between a blindfolded human and a language-only AI model. We supplemented STAR situations with textual descriptions written by us for this purpose. Both participants then interacted with Video-LLaMa 7B via \href{https://bit.ly/VLLaMa}{the Hugging Face Web demo}. Participants were instructed to always set the temperature parameter to the minimum value (0.1) to increase the model output consistency across trials. Two prompts were used: one included only the video attachment and question, while the other added the textual description (Figure \ref{fig:collab_game}). This setup enabled analysis of responses across multimodal and text-only inputs. We also encouraged players to modify the given prompts after the first interaction to test different prompting strategies. Participants recorded their observations on \href{https://miro.com/index/}{a digital Miro board}. %\footnote{\url{}}.    

%\begin{figure*}[t]
%    \centering
%    \includegraphics[width=\textwidth]{images/collaborative_game.pdf} 
%    \caption{Structure of the collaborative game. \textbf{Part 1:} players interact in pairs to solve different visual reasoning tasks. Different prompts are provided to compare responses from the blindfolded player, the LLM, and the VLM. \textbf{Part 2:} Mock-ups under evaluation. }
%    \label{fig:collab_game}
%\end{figure*}

\paragraph{Part 2: Mock-up evaluation.} In the second phase, participants reviewed graphical mock-ups of a proposed Web App designed to support a future extended study (Figure \ref{fig:mockup}). %We used \href{https://www.mentimeter.com/}{Mentimeter} for collecting feedback in this phase.  %\footnote{\url{https://www.mentimeter.com/}}  
Participants ranked features by importance (Figure \ref{fig:rank}), explained their rankings in free-text responses, and answered more detailed questions on individual features, through \href{https://www.mentimeter.com/}{Mentimeter}. Readers can refer to the supplementary material for the complete question and rating structure used for this phase. Based on the survey findings, a key focus was evaluating whether a scene graph representation of the video was perceived as a useful addition to textual responses.
%, and whether text and graph outputs should be displayed side-by-side or on separate pages.   

%\begin{figure*}[t]
%    \centering
%    \includegraphics[scale=0.5,  trim={10 60 30 10}, clip]%{images/mockups.pdf} 
%    \caption{Mock-ups under evaluation. }
%    \label{fig:mockup}
%\end{figure*}

\subsection{Participants}
We held the workshop in December 2024 with 8 participants, organized into four teams for the collaborative game. To explore how trust develops in early interactions with a new model and data modality, specifically videos (\textbf{RQ1}), we recruited participants with varying familiarity with AI chatbots and LLMs but no significant experience with VLMs. Simultaneously, to gather feedback for designing the Web App (\textbf{RQ2}), we searched for participants with digital design and development skills. 
Participants were recruited at the institution of one of the paper authors via a team collaboration platform among students and faculty members both well-versed in theoretical Software Engineering (SE) and Digital Design and Innovation (DDI)  aspects and actively engaged in real-world product design. 
Specifically, five students in SW and DDI, two Professors in DDI, and one researcher in participatory design took part, all proficient English speakers. Participants were fully informed about the study purpose and procedures through a detailed information sheet and formal consent form, which clarified that photos, group discussions, and activity outcomes would be recorded for academic purposes while maintaining confidentiality and anonymity of the provided data. The full consent form and information sheet is enclosed in the supplementary. Participation was voluntary, with the option to withdraw at any time; all participants completed the session.  

\subsection{Results}
\paragraph{Collaborative game.} After each task, players (1) compared the answers of the blindfolded partner with ChatGPT's responses (2) analysed Video-LLaMa's answers. We analysed player notes on Miro and re-watched the full workshop discussion recorded through a videoconferencing tool to summarise the results. 
For (1), the blindfolded player and GPT4-o achieved the same average ratio of correct answers across question types, i.e., 75\% accuracy. 
Interestingly, in some cases, the LLM could guess the correct answer while the blindfolded player could not, particularly for perception-level questions, where GPT reached 100\% accuracy, surpassing the 91.67\% achieved by humans. 
Participants felt that the verbs chosen in the video description and questions (e.g., "throwing" vs. "putting down") were especially impactful on the players' understanding, leading some teams to test different verb synonyms in prompts. In temporal reasoning questions, GPT accuracy dropped below the blindfolded players’ to 75\%, below the blindfolded players’ 83.33\%. 
As expected, predictive questions requiring causal reasoning and meta-learning proved hardest for both the LLM and participants, with both averaging 50\% accuracy. \\ Notably, in one case, both the human and GPT, despite providing the wrong answer, pointed out a discrepancy in object naming between the video caption and question that hindered comprehension. \\ 
For (2), querying Video-LLaMa with multimodal prompts led to the lowest accuracy for all question types: 29.63\% overall, 25\% on perception-level questions and 33.33\% on temporal reasoning questions. 
The model frequently hallucinated objects not present in the video, providing contradictory (``It said \texttt{[}the plate\texttt{]} was on the table at the end but also that the person was still holding it.") 
and inconclusive answers (kept looping over the sentence ``the person is standing in a kitchen, and there is a white door in the background..." without converging). 
Many noted that the VLM often avoided directly answering questions, regardless of the question asked, even when providing generally accurate descriptions of the video (``It didn't answer the question, it just described the video."). 
Differently from GPT, the VLM relative performance was higher on temporal than on perception-level questions, likely due to being fine-tuned on temporally ordered frames. Predictive questions were again the hardest, with only 8.33\% accuracy. 
though Video-LLaMa generally provided more conservative answers than GPT, especially on forecasting (``The model said it is not possible to determine what the person would eat", the model replied ``We cannot determine whether they closed anything after holding the food because the action itself is not seen in the video"). 
Participants felt that GPT provided ``more believable answers", even when incorrect, compared to Video-LLaMa. Interestingly, one team experimented with using the VLM in text-only mode despite not being instructed to do so. While the model still struggled to answer correctly, adding textual context occasionally improved its guesses, reflecting the same language bias reported in related work \cite{gou2024eyes,huai_debiased_2024}. %(Section \ref{sec:lit_mapping}).  %the VLM's tendency to prioritise text over visual modalities \cite{gou2024eyes,huai_debiased_2024}. \\

For future iterations of the study, we wrapped up the session asking participants \textit{What would you change about this game and why?} Participants expressed distrust in Video-LLaMa's capabilities, suggesting it be removed. They felt the game structure prioritised task completion over prompt refinement and successive model interactions and recommended rephrasing prompts to allow for diverse verb forms. They also enjoyed the blindfold component, confirming that this strategy encouraged reliance on senses beyond vision. 
%At the end of the study tasks, participants reflected on their findings in a brief debriefing session. % Overall, prediction questions were deemed the hardest to answer, even for blindfolded players. Participants highlighted the significant impact of verb choices in video captions and prompts (e.g., “throwing” vs. “putting down”) on understanding. 
%For this reason, some teams tested different verb synonyms in prompts. 
%\vspace{-1em}
\paragraph{Mock-up evaluation.} Participants' feature rankings are summarised in Figure \ref{fig:rank}. For \textbf{Page 1}, most users prioritised basic features over trust metric trackers like the TrustMeter, i.e., a feature we took from Mehrotra et al., \cite{mehrotra2024integrity} that allows users to set a score from 0 (distrust) to 100 (complete trust) to the model. Users preferred interacting with the model first before assigning a trust score and deemed these features less intuitive, especially for AI novices. 
They also suggested replacing continuous likelihood sliders with categorical ratings such as Likert-scale radio buttons. 
For \textbf{Page 2}, participants were presented with two app fields, one for evaluating the VLM's textual response and the other for evaluating the model graph-based response, i.e., the VLM output when asked to generate, in addition to the textual answer, a graph representing the observed scene. They rated textual and graph-based responses from the VLM as equally important, with 71\% preferring to display both on a single page.  
They suggested improvements like adding a scrolling sidebar, making the graph interactive (e.g., draggable, zoomable), and using colours to highlight node/edge types and importance. To enhance usability, they recommended explanatory pop-ups and removing unnecessary elements like the ``Response" divider.
For \textbf{Page 3}, participants ranked the model text rating and the return button to revisit the responses highest (Figure \ref{fig:rank}). 
The TrustMeter was deemed more relevant on \textbf{Page 3} than on Page 1, while the graph rating and free-text feedback were seen as less important. Specifically, participants preferred annotating individual graph nodes/edges over rating the entire graph. While users felt trust scores were more likely to be completed than free-text input, they suggested requiring textual justifications in the case of extreme TrustMeter scores. They also suggested numbering rating questions for clarity. 
Overall, users found the app moderately intuitive and tasks easy to understand but noted that content balance on each page could be improved. While they were neutral about future use as they perceived no benefit beyond this research context, they felt the app could effectively support TAI studies and collaborative gamified evaluations. %Overall, users reported was averagely intuitive and the tasks easy to understand, though content balance in each page could be improved. They expressed neutrality towards using the app in the future. While they saw no personal benefit in reusing the app for repeated VLM interaction unless more context was provided, they also felt the Web App could adequately support a study designed like the collaborative game they just played, benefitting TAI studies and making the evaluation process smoothers for users.  % (see supplementary material).   
%Feedback on the web app design. \\
%with links to graphics showing the GUI mock-ups and Mentimeter results
 
\section{Open challenges and opportunities} 
%Our proposed taxonomy to categorize user-VLM trust research enabled us to evaluate the current coverage of key factors contributing to trust building. Recognizing the scarcity of studies directly involving users, we organized a workshop with potential VLM users from a digital design and innovation background to distill preliminary requirements for future AI trust research. 
Insights from the literature review and workshop revealed key research challenges and opportunities for improvement. 
\paragraph{Domain complementarity.} Fewer than 28\% of the retrieved papers on trust and TAI keywords focus on Computer Vision and Vision Language Models. While our review emphasizes recent work and may exclude earlier studies, it reveals a gap: TAI research involving users often focuses on language modalities, whereas visual perception studies prioritize performance metrics over trust properties. This complementarity of focus highlights the need for multidisciplinary collaboration across technical fields, Philosophy, Cognitive Science, and Digital Design to advance AI trust research. The need for multidisciplinarity is also underscored by the varying ethical implications across domains. For example, studies on autonomous driving in this survey focus narrowly on model hallucinations, overlooking deeper ethical considerations in designing for high-risk applications. 
\paragraph{Beyond hallucinations and adversarial learning.} As anticipated in the previous paragraph, research on VLM trust largely focuses on reducing hallucinations and countering adversarial attacks, with benchmarks reflecting similar trends and limited diversity, despite providing a relatively higher coverage of cognitive capabilities. 
Other integrity issues already examined for LLMs \cite{huang2024trustllm}, such as preference bias, data leakage, accountability bugs, emotional awareness, ethical dilemmas, decision opacity, and rhetorical manipulation, require more attention in VLM studies. In addition, advanced reasoning challenges like causality and meta-learning, also evident from the workshop results, remain under-explored. 
\paragraph{Bridging vision and language with graphs.} Scene graphs offer a promising way to abstract concepts creatively from visual components and structure the model outputs and reasoning chains. Moreover, to address VLMs' over-reliance on language, exploring a hybrid modality, which is decomposable into text triples yet grounded in compositional visual elements, opens up interesting opportunities of investigation.   
\paragraph{Human-in-the-loop or Model-in-the-loop?} VLM trust and trustworthiness studies rarely involve users directly. Even in aligning models with human values, feedback and labeling are often automated using LLMs. While these models can act as valuable collaboration aids (e.g., for brainstorming), user engagement is essential to capture trust requirements. Drawing from lessons learned during our workshop, we propose preliminary requirements for future VLM trust studies.

\paragraph{Requirements for user-VLM trust studies.} The exploratory workshop allowed us to identify a few pre-requisites for conducting larger scale studies with users: 
\begin{itemize}
    \item{One core pre-requisite emerging from the workshop is prioritising user agency over task completion, allowing multi-turn interactions and prompt refinement.}
    \item{Trust metrics should be contextualised through clear explanations and concrete research objectives to foster engagement and motivate users to contribute to broader impacts.}
    \item{Studies should explore different graphical representations, in addition to text and visuals, to foster user engagement. In particular, graph representations hold potential for enhancing user interaction and engagement, as these representational format allows researchers to collect targeted feedback on individual components (relationship nodes and edges).}
    \item{As trust can drop sharply after initial failures (see the case of Video-LLaMa), the continuous tracking of trust metrics throughout the evaluation tasks is essential. However, trust scores can be hard to assess a priori and ice-breaking interaction rounds could help users calibrate their TrustMeter.}
    \item{Ensuring to involve a representative and diversified user base is essential for capturing multi-disciplinary voices, especially those from under-represented groups that are particularly vulnerable to the implicit bias embedded in AI models.}
\end{itemize}
    %Diversity of participant base: stratified across familiarity with the technology, discipline/background to foster multidisciplinarity and inclusive of underrepresented groups to spot implicit model biases. 

 These research trajectories and requirements can guide researchers, practitioners, and policymakers in advancing the study of user-VLM trust, producing key societal impact. They offer a foundation for larger-scale, targeted user studies aimed at distilling design principles for specific applications, ultimately helping bridge the gap between regulation and real-world deployment.   

 \begin{ack}Redacted for double-blind review.
\end{ack}
\bibliography{ecai25}

%\clearpage
%\pagebreak
\input{supplement.tex}

\end{document}

%% file: supplement.tex
\section*{Supplementary material}
In the following sections, we provide additional details on the structure of the workshop conducted with participants from a Design and Development background to collect preliminary requirements for carrying out larger scale studies on user-VLM trust, a theme still largely unexplored in the related literature. Specifically, we provide details on the tasks and question types characterising each phase of the workshop, the collaborative game and the Web app mock-up evaluation. Lastly, we provide the forms we used to collect participants' consent, though opportunely redacted for anonimity. 

\subsection*{Workshop part 1: collaborative game}

\begin{figure*}[t]
    \centering
    %{\textwidth} 
        \centering
    \includegraphics[width=0.9\textwidth,  trim={0 50 50 50}, clip]{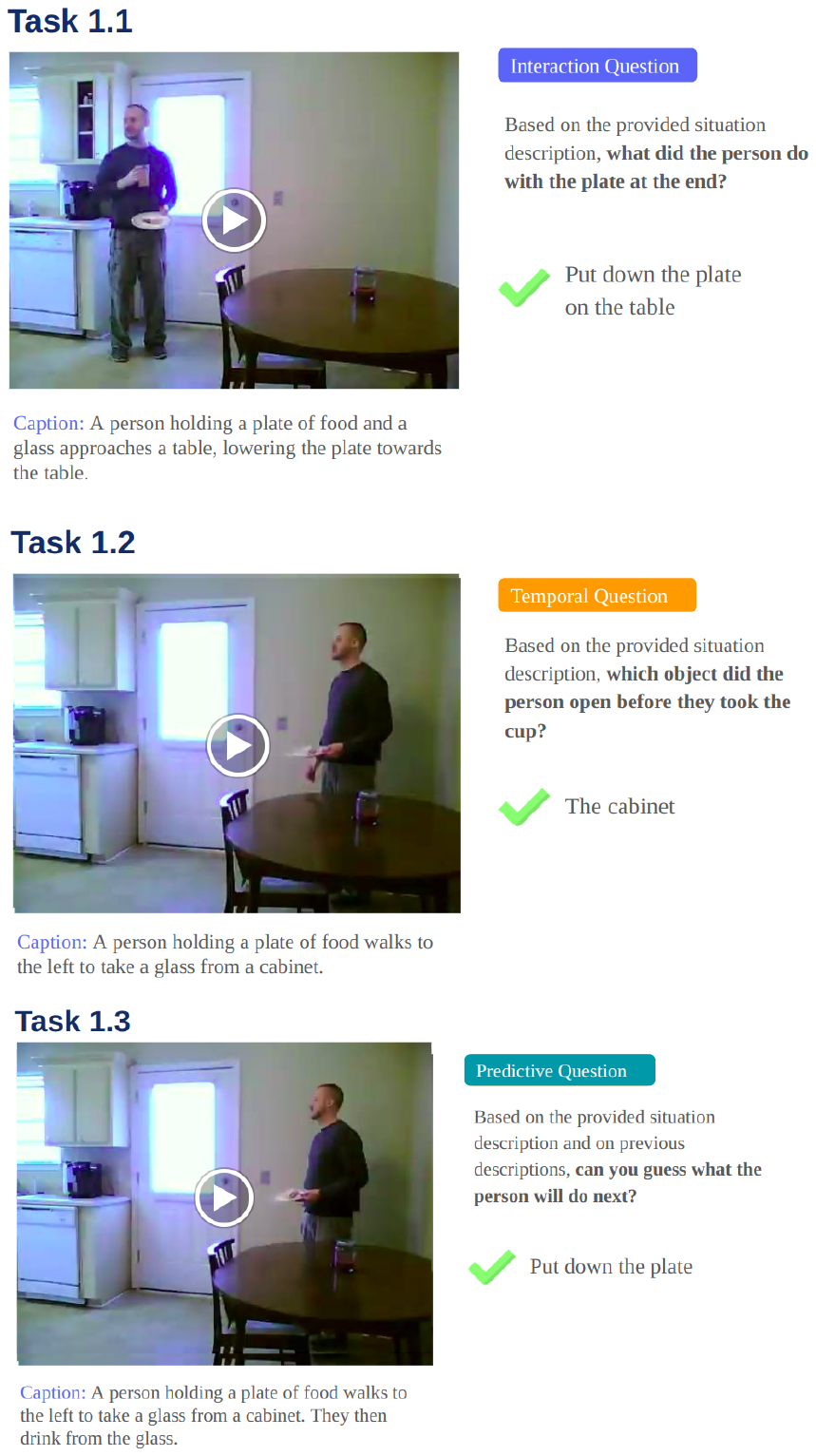} 
    \caption{\textbf{Collaborative Game Task 1}}
    \label{fig:task1}
    %\vspace*{3ex}
\end{figure*}

\begin{figure*}[t]
    \centering
    %{\textwidth} 
        \centering
    \includegraphics[width=0.9\textwidth,  trim={0 50 50 50},clip]{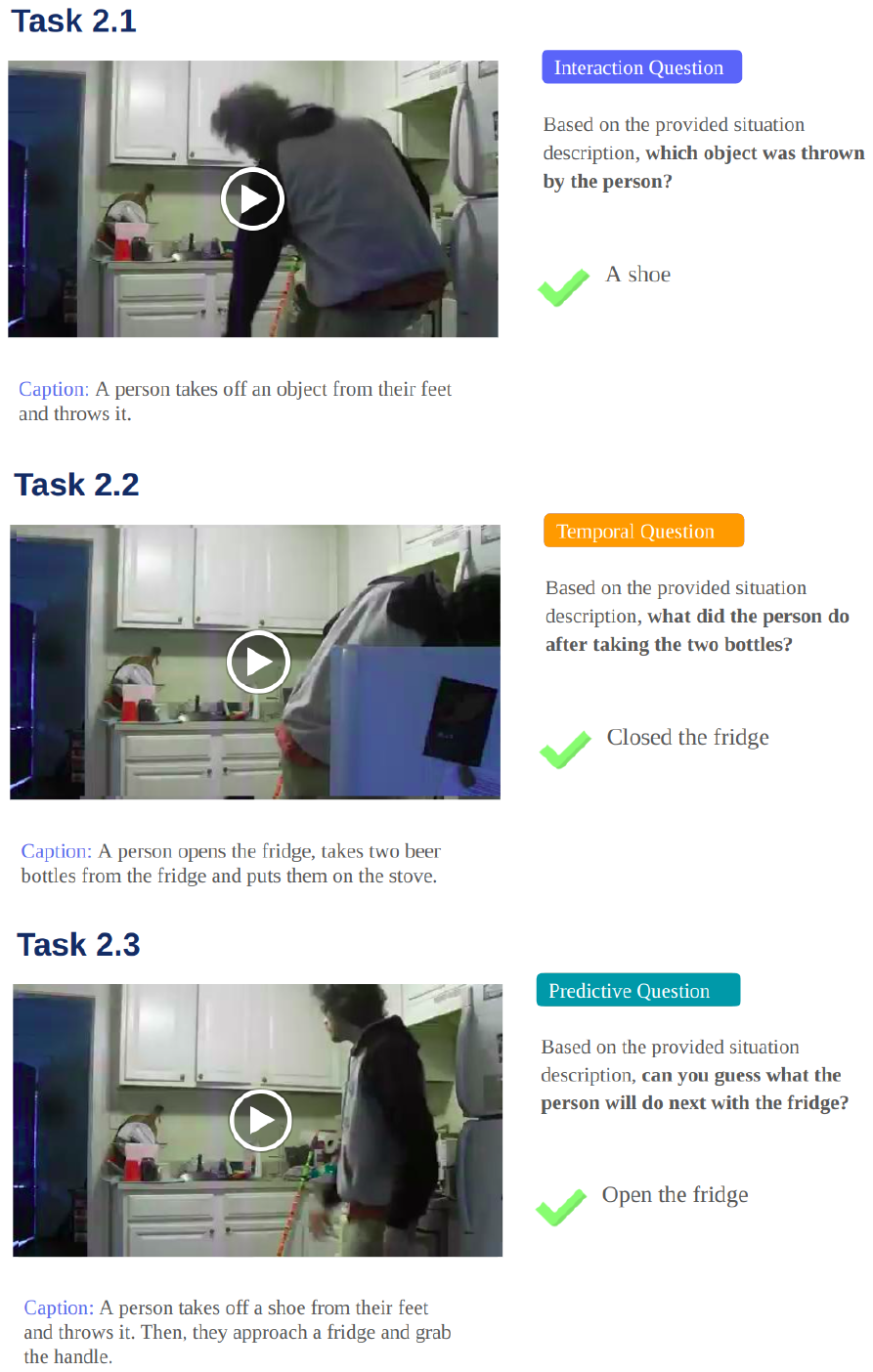} 
    \caption{\textbf{Collaborative Game Task 2}}
    \label{fig:task2}
    \vspace*{3ex}
\end{figure*}

\begin{figure*}[t]
    \centering
    %{\textwidth} 
        \centering
    \includegraphics[width=0.85\textwidth,  trim={0 50 50 50},clip]{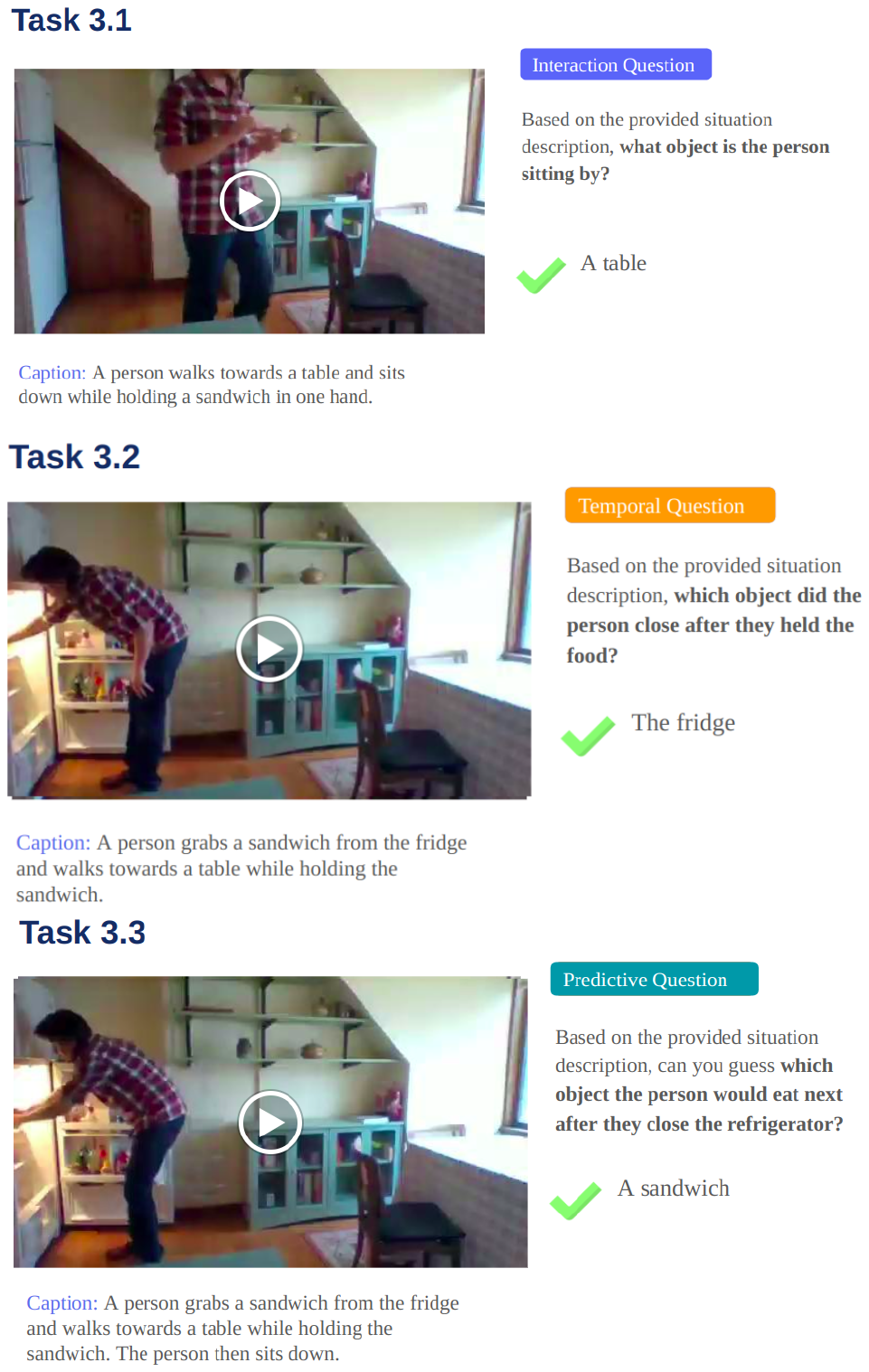} 
    \caption{\textbf{Collaborative Game Task 3}}
    \label{fig:task3}
    \vspace*{3ex}
\end{figure*}

\begin{figure*}[t]
    \centering
    %{\textwidth} 
        \centering
    \includegraphics[width=0.9\textwidth,  trim={0 50 0 0},clip]{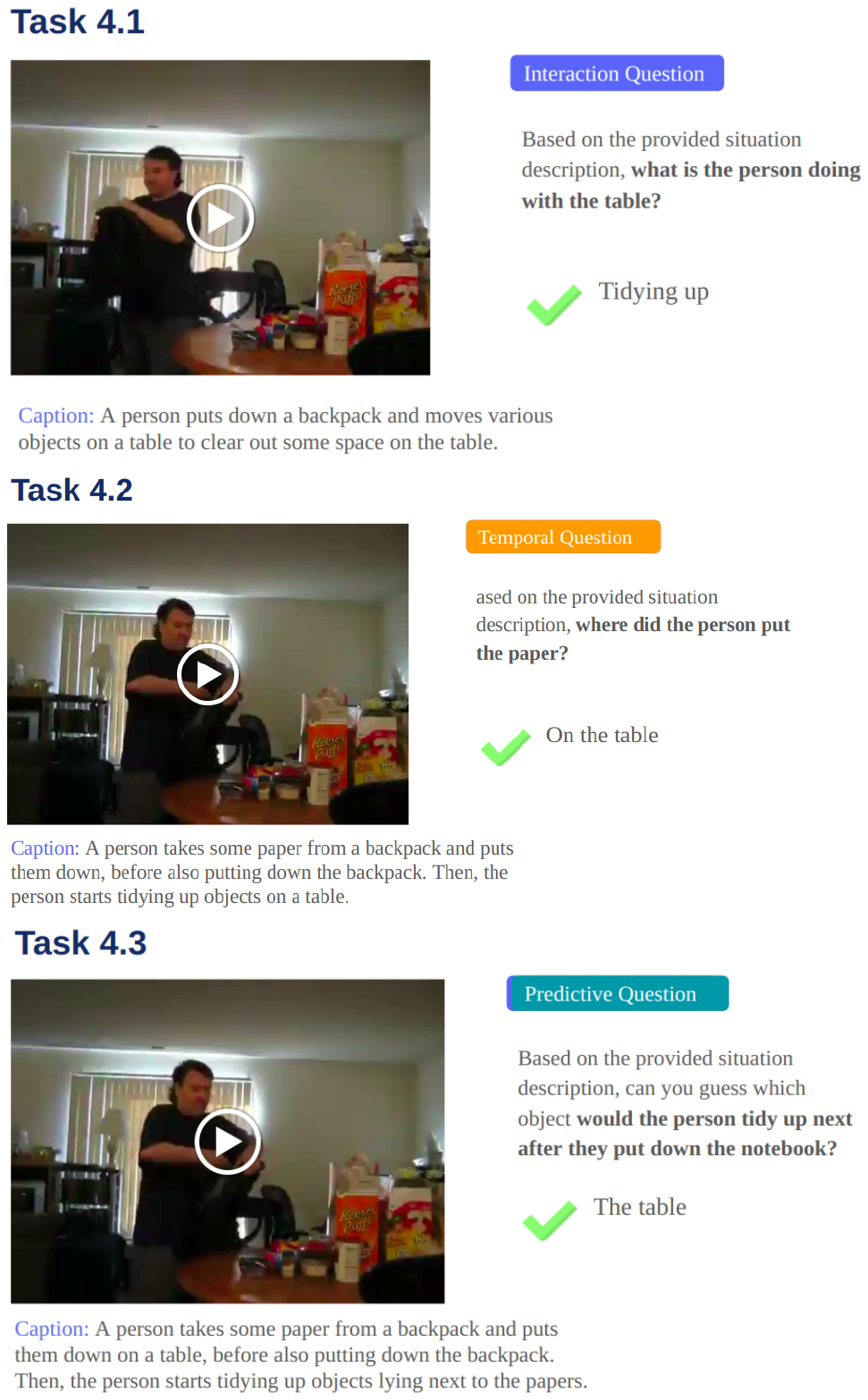} 
    \caption{\textbf{Collaborative Game Task 4}}
    \label{fig:task4}
    \vspace*{3ex}
\end{figure*}

In the first phase of the workshop, participants were paired up and presented with five tasks, one of which served as ice-breaking round. Tasks and questions are shown in Figures \ref{fig:task1} through \ref{fig:task4}. Each task consisted in watching a video and answering a series of questions. 

Videos were taken from the Situated Reasoning in Real-world Videos (STAR) dataset \cite{wu2021star}, a large-scale collection of videoclips built on top of the Charades benchmark \cite{sigurdsson2017charades} and depicting everyday human interaction. We took the original multiple-choice questions from STAR and rephrased them as open-ended questions to increase the complexity of the test. We also complemented the STAR questions with hand-written captions to provide a textual input to feed the LLM and VLM models. 

In particular, we focused on : i) interaction, ii) temporal, and iii) predictive questions. 
\textit{Interaction questions} serve as a fundamental test for understanding how humans interact with objects within a given scenario. 
\textit{Sequence, or temporal questions} are conceived to assess a system’s ability to reason about temporal relationships by identifying the order of events in dynamic contexts. \textit{Predictive questions} focus on predicting plausible future actions based on the current situation. Only the initial portion (one-quarter) of an action sequence is shown, while the remainder is hidden to test anticipatory reasoning. 

We chose STAR among the datasets surveyed in Section 3.3 as, notably, this dataset also provides frame-by-frame scene graph annotations of the situations depicted in each video. Thus, it can facilitate future interaction studies examining graphs as a hybrid modality between visuals and text VLM model responses. 

\begin{figure*}[t]
    \centering
    %{\textwidth} 
        \centering
    \includegraphics[width=\textwidth,  trim={0 0 0 180}, clip]{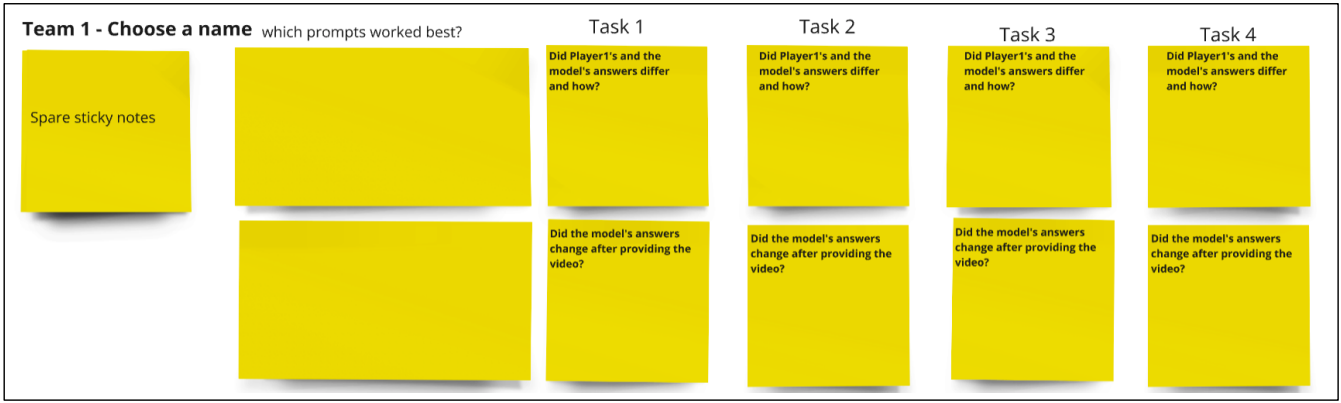} 
    \caption{Question structure in the Miro digital board.}
    \label{fig:miro}
    \vspace*{3ex}
\end{figure*}

After completing each task, participant pairs were asked to answer a set of pre-determined questions on a digital Miro board, as illustrated in Figure \ref{fig:miro}. We used team answers on the Miro board as well as group discussions recordings collected through a videoconferencing tool to code the workshop results reported in the paper. 

\subsection*{Workshop part 2: Mock-up evaluation}
\begin{figure*}[t]
    \centering
    %{\textwidth} 
        \centering
    \includegraphics[width=0.85\textwidth,  trim={0 50 50 50},clip]{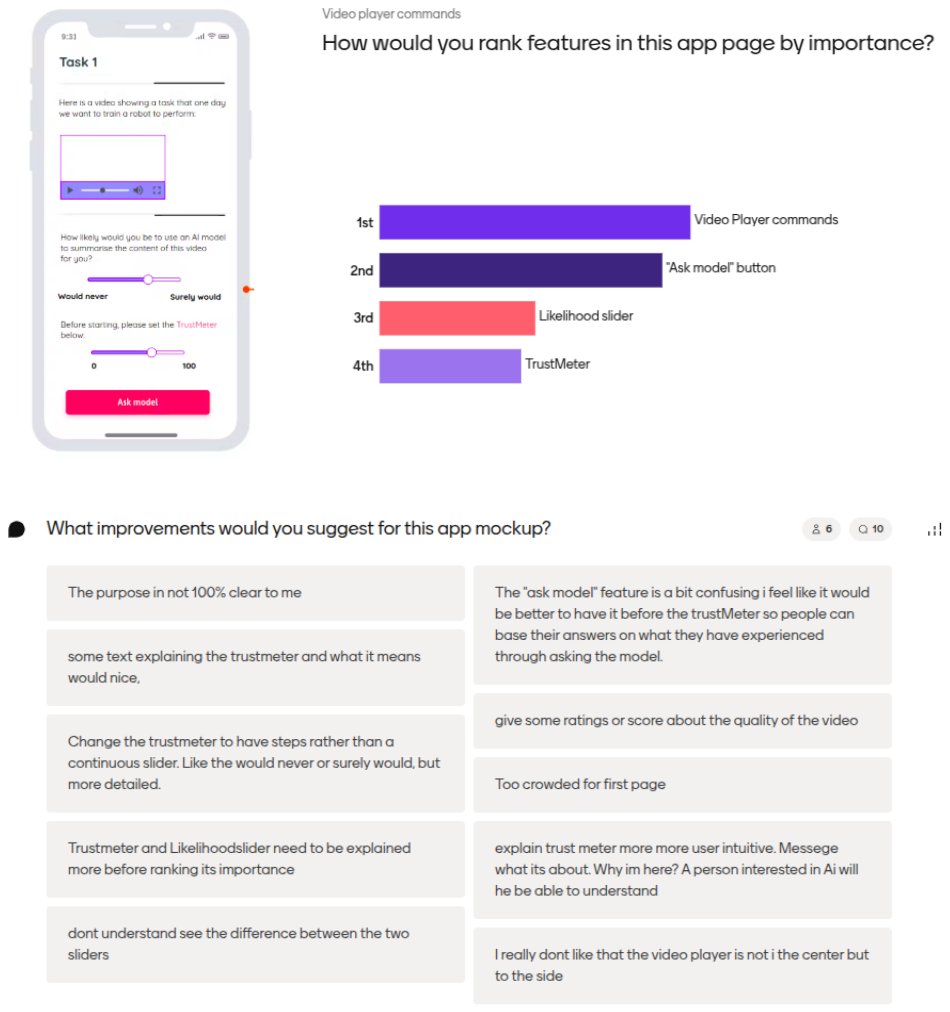} 
    \caption{\textbf{Feedback on mock-up page 1.}}
    \label{fig:menti1}
    \vspace*{3ex}
\end{figure*}

\begin{figure*}[t]
    \centering
    %{\textwidth} 
        \centering
    \includegraphics[width=0.85\textwidth,  trim={0 50 50 50},clip]{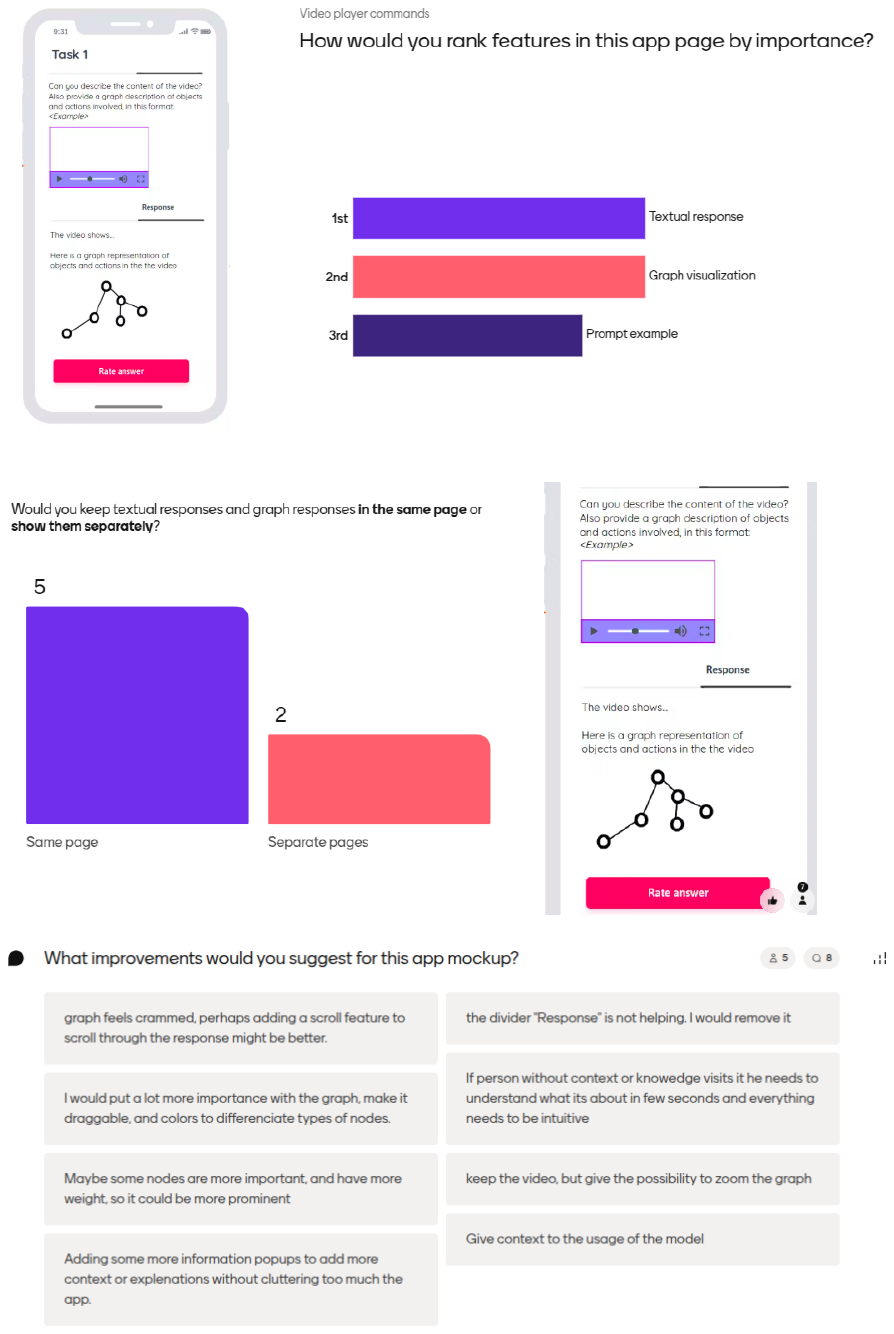} 
    \caption{\textbf{Feedback on mock-up page 2.}}
    \label{fig:menti2}
    \vspace*{3ex}
\end{figure*}

\begin{figure*}[t]
    \centering
    %{\textwidth} 
        \centering
    \includegraphics[width=0.85\textwidth,  trim={0 50 50 50},clip]{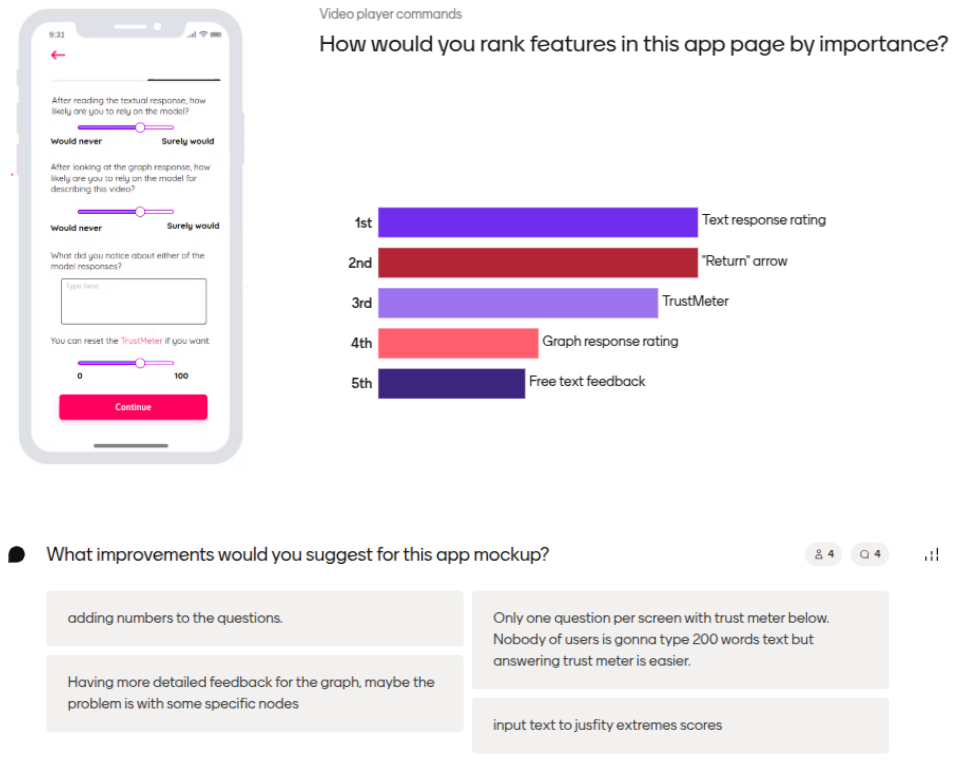} 
    \caption{\textbf{Feedback on mock-up page 3.}}
    \label{fig:menti3}
    \vspace*{3ex}
\end{figure*}

\begin{figure*}[t]
    \centering
    %{\textwidth} 
        \centering
    \includegraphics[width=0.85\textwidth,  trim={0 50 50 50},clip]{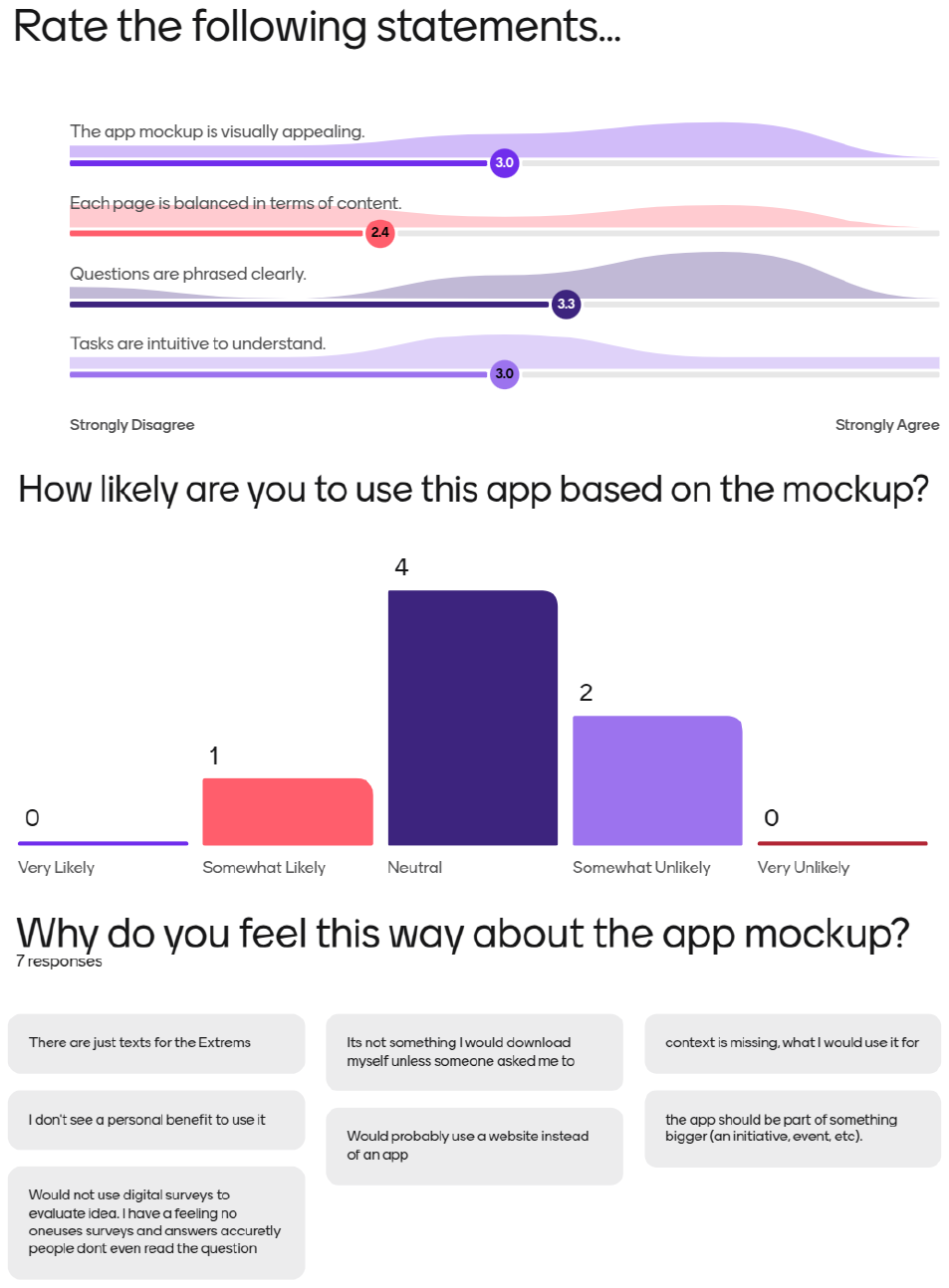} 
    \caption{\textbf{Final user ratings.}}
    \label{fig:menti4}
    \vspace*{3ex}
\end{figure*}

\begin{figure*}[t]
    \centering
    %{\textwidth} 
        \centering
    \includegraphics[width=0.85\textwidth,  trim={0 50 50 50},clip]{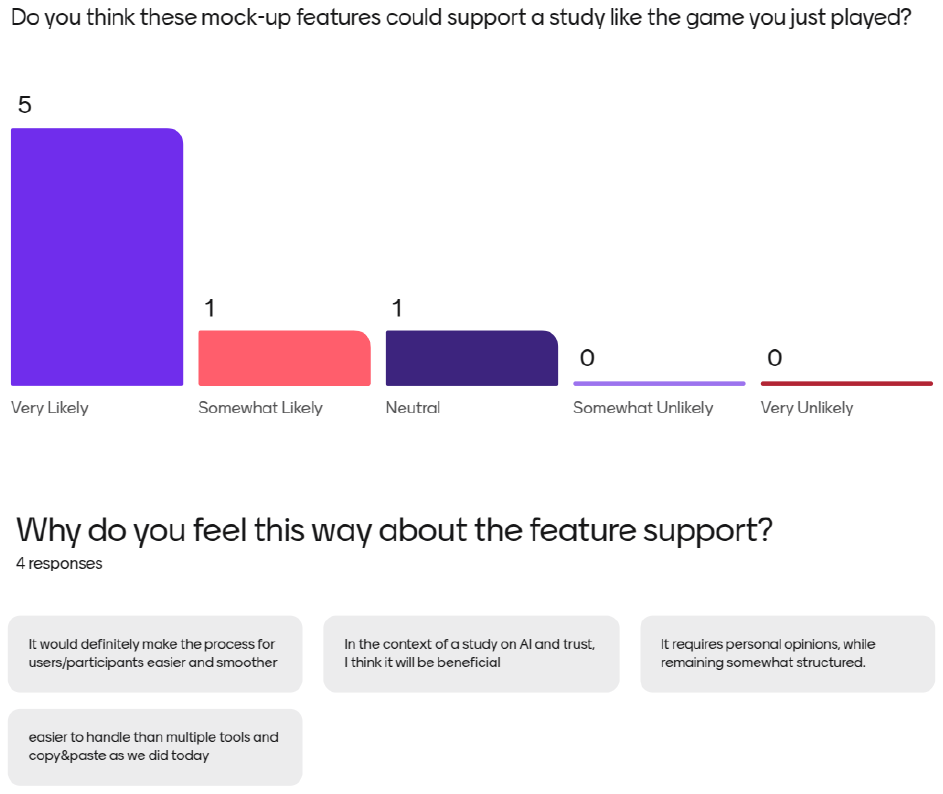} 
    \caption{\textbf{Final user ratings}}
    \label{fig:menti5}
    \vspace*{3ex}
\end{figure*}

In the second part of the workshop, we collected participants' feedback through Mentimeter. The questions and responses collected in this phase are reported in Figures \ref{fig:menti1} through \ref{fig:menti5}.

\subsection*{User Consent Form and Information Sheet}
\begin{figure*}[t]
    \centering
    %{\textwidth} 
        \centering
    \includegraphics[width=0.9\textwidth, trim={0 0 0 0}, clip]{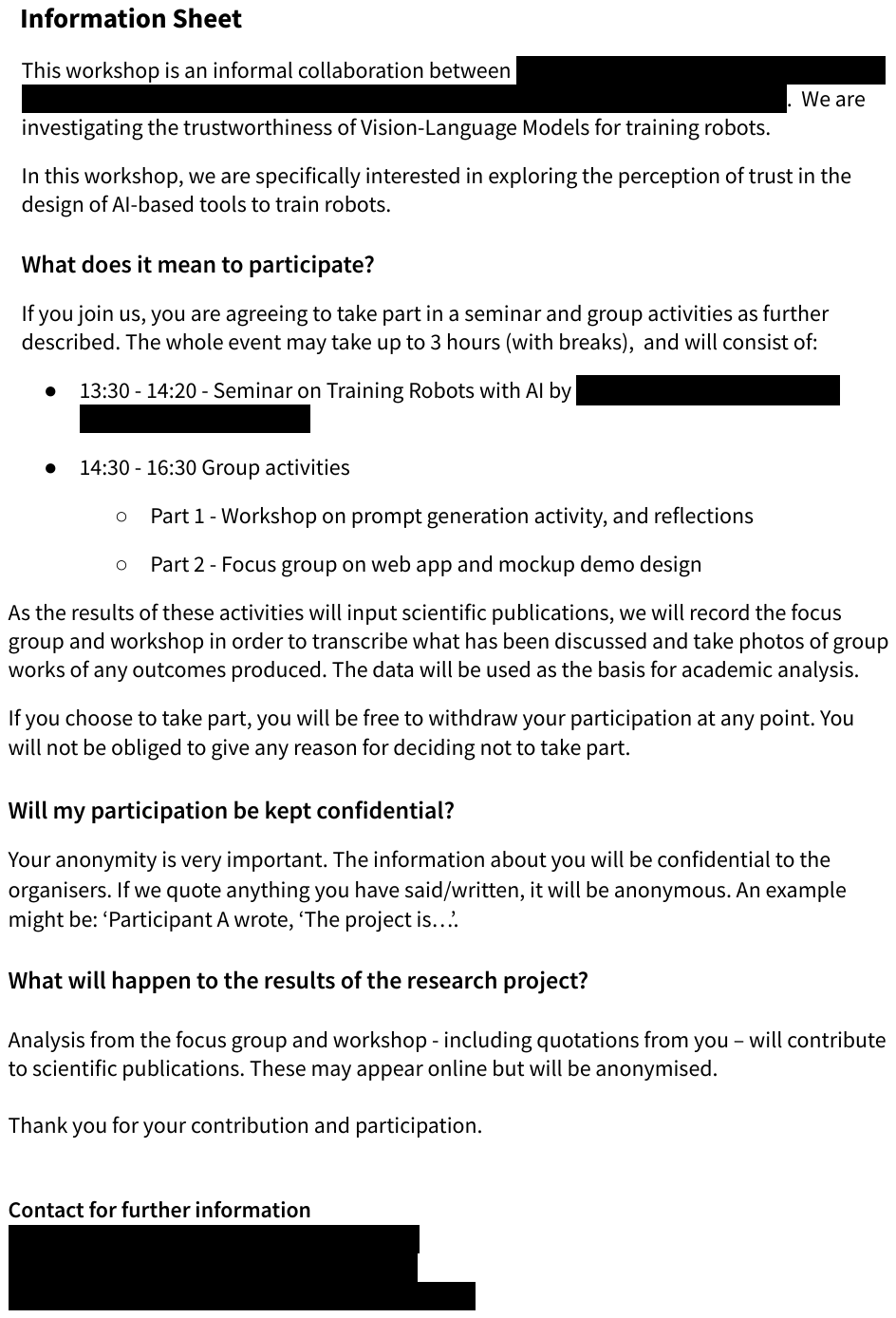} 
    \caption{\textbf{Participants' information sheet}}
    \label{fig:sheet}
    \vspace*{3ex}
\end{figure*}

\begin{figure*}[t]
    \centering
    %{\textwidth} 
        \centering
    \includegraphics[width=0.9\textwidth, trim={0 0 0 0}, clip]{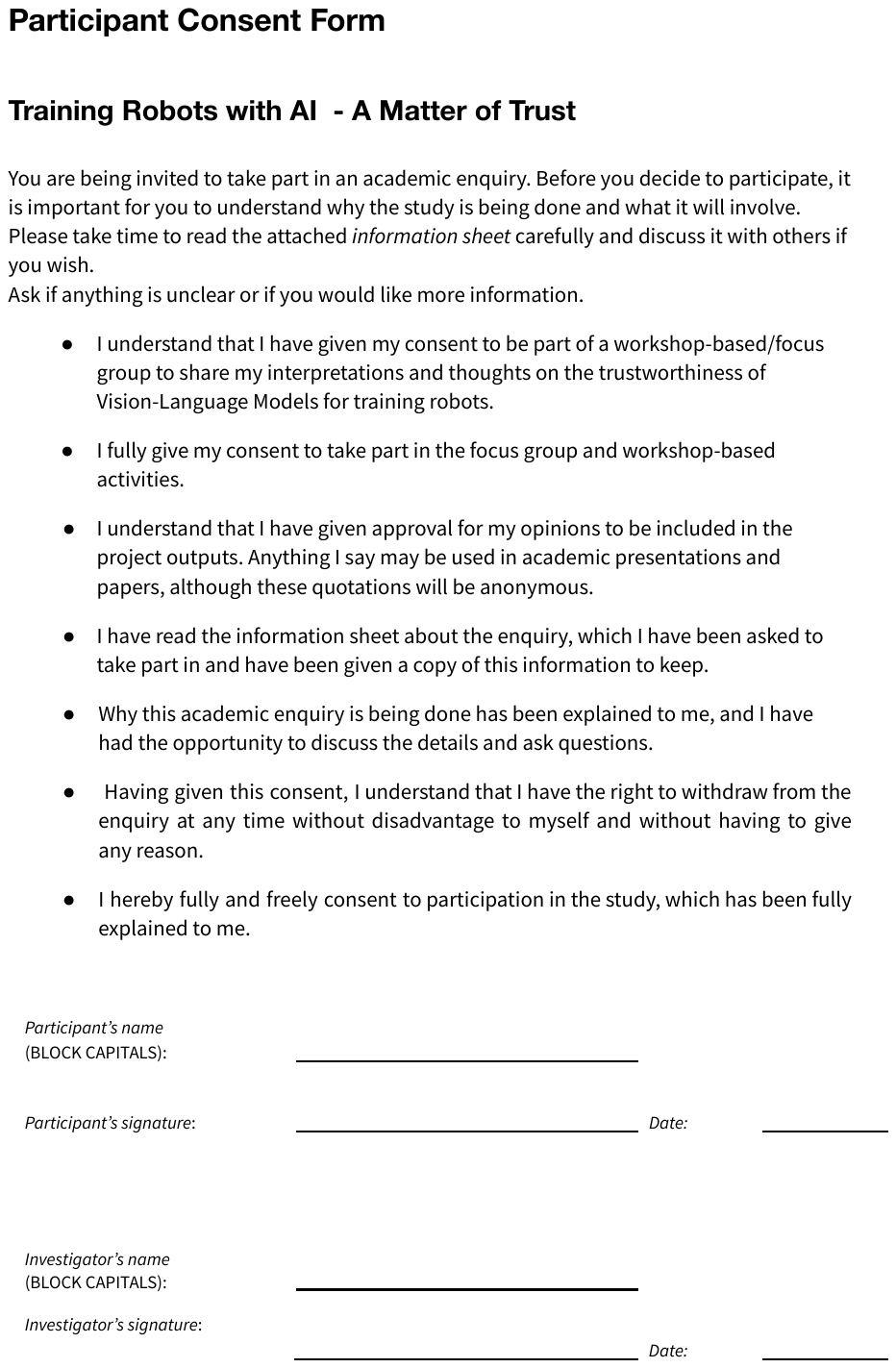} 
    \caption{\textbf{Participants' consent form}}
    \label{fig:consent}
    \vspace*{3ex}
\end{figure*}

Figures \ref{fig:sheet} and \ref{fig:consent} shows the user consent form and information sheet, anonimised for double-blind review purposes. 

    \begin{figure}[H]
    \centering
    %{\textwidth} 
        \centering
    \includegraphics[width=0.4\textwidth,  trim={0 220 350 0}, clip]{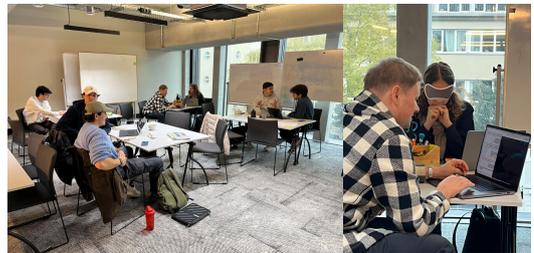} 
    \caption{\textbf{Design and Development workshop:} players interact in pairs using different prompts to compare responses from the blindfolded player, the LLM, and the VLM.}
    \label{fig:workshop}
    \vspace*{3ex}
\end{figure}

\begin{figure*}[b]
    \centering
    %{\textwidth} 
        \centering
    \includegraphics[width=\textwidth, trim={0 0 0 0}, clip]{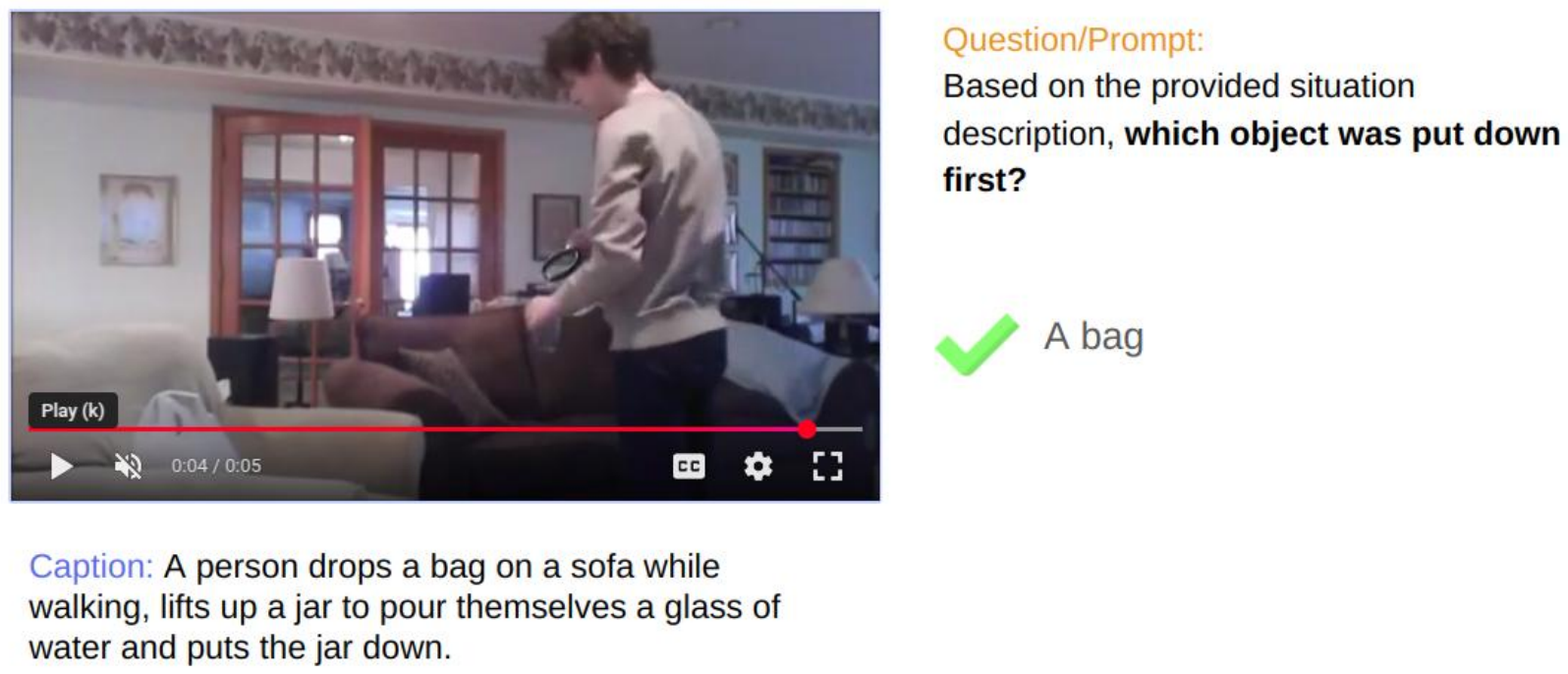} 
    \caption{\textbf{Collaborative Game Trial Round}}
    \label{fig:trial}
    \vspace*{3ex}
\end{figure*}

%%%Mentimeter

%%%final forms

